\documentclass{article}

\usepackage{microtype}
\usepackage{graphicx}
\usepackage{subcaption}
\usepackage{booktabs} 

\usepackage{hyperref}

\usepackage[preprint]{icml2026}

\usepackage{caption}
\usepackage{amsmath}
\usepackage{amssymb}
\usepackage{mathtools}
\usepackage{amsthm}
\usepackage{multirow} 

\usepackage[capitalize,noabbrev]{cleveref}

\theoremstyle{plain}

\theoremstyle{definition}

\theoremstyle{remark}

\usepackage[textsize=tiny]{todonotes}

\icmltitlerunning{Analyzing Physical Adversarial Example Threats to Machine Learning in Election Systems}

\begin{document}

\twocolumn[
  \icmltitle{Analyzing Physical Adversarial Example Threats to Machine Learning in Election Systems}



  \icmlsetsymbol{equal}{*}


  \begin{icmlauthorlist}
    \icmlauthor{Khaleque Md Aashiq Kamal}{zzz}
    \icmlauthor{Surya Eada}{}
    \icmlauthor{Aayushi Verma}{xxx}
    \icmlauthor{Subek Acharya}{yyy}
    \icmlauthor{Adrian Yemin}{xxx}
    \icmlauthor{Benjamin Fuller}{xxx}
    \icmlauthor{Kaleel Mahmood}{yyy}
  \end{icmlauthorlist}
  \icmlaffiliation{zzz}{Department of Electrical, Computer and Biomedical Engineering, University of Rhode Island, Kingston, RI, USA}
  \icmlaffiliation{yyy}{Department of Computer Science and Statistics, University of Rhode Island, Kingston, RI, USA}
  \icmlaffiliation{xxx}{Voting Technology Center, University of Connecticut, Storrs, CT, USA}

  \icmlcorrespondingauthor{Khaleque, Md Aashiq Kamal}{aashiqkamal@uri.edu}

  \icmlkeywords{Machine Learning, ICML}

  \vskip 0.3in
]



\printAffiliationsAndNotice{}  

\begin{abstract}
Developments in the machine learning voting domain have shown both promising results and risks. Trained models perform well on ballot classification tasks ($>99\%$ accuracy) but are at risk from adversarial example attacks that cause misclassifications. In this paper, we analyze an attacker who seeks to deploy adversarial examples against machine learning ballot classifiers to compromise a U.S. election. We first derive a probabilistic framework for determining the number of adversarial example ballots that must be printed to flip an election, in terms of the probability of each candidate winning and the total number of ballots cast. Second, it is an open question as to which type of adversarial example is most effective when physically printed in the voting domain. We analyze six different types of adversarial example attacks: $l_{\infty}$-APGD, $l_{2}$-APGD, $l_{1}$-APGD, $l_{0}$ PGD, $l_{0}+l_{\infty}$ PGD and $l_{0}+\sigma$-map PGD. Our experiments include physical realizations of $144,000$ adversarial examples through printing and scanning with four different machine learning models.  We empirically demonstrate an analysis gap between the physical and digital domains, wherein attacks most effective in the digital domain ($l_{2}$ and $l_{\infty}$) differ from those most effective in the physical domain ($l_{1}$ and $l_{2}$, depending on the model). By unifying a probabilistic election framework with digital and physical adversarial example evaluations, we move beyond prior close race analyses to explicitly quantify \textit{when} and \textit{how} adversarial ballot manipulation could alter outcomes. \noindent\textbf{Code available at:} \url{https://github.com/aashiqkamal/VotingThreats}
\end{abstract}
\section{Introduction}

Elections in the United States are conducted at the state level, primarily through the use of paper ballots. As of 2026, the \href{https://verifiedvoting.org/verifier/#mode/navigate/map/voteEquip/mapType/ppEquip/year/2026}{Verified Voting Database} states that $69.4\%$ of elections in the United States used hand marked paper ballots. This number is increasing with the voter-verifiable paper trail being the gold standard for election integrity~\cite{national2018securing}. In determining election results, the main task associated with marked paper ballots is to determine whether each bubble on the ballot is blank or filled.\footnote{There are some important secondary tasks including locating the bubbles, which is done through a combination of an election definition, timing marks, alignment, and segmentation~\cite{kiayias2007authentication}.} This binary classification task can be accomplished by an optical scanner, or alternatively by using a machine learning model that has been trained on ballot images~\cite{CNNBallot,Busting}. It is important to note that machine learning models are vulnerable to adversarial examples~\cite{FGSMpaper}. An adversarial example is an input $x$ that has imperceivable noise $\delta$ added such that human beings can recognize the input, but machine learning models misclassify $x+\delta$. In~\cite{Busting}, it was shown that machine learning models trained for ballot classification were vulnerable to adversarial examples that were physically printed and scanned. However, the attacks in~\cite{Busting} were limited to creating imperceivable adversarial examples using only the $l_\infty$ norm. There are multiple different norm attacks that can create imperceivable adversarial examples~\cite{AttackL1, AttackL0}. Furthermore, it is important to consider different norm attacks because robustness to one type of norm attack does not guarantee robustness to other norm attacks~\cite{Union, RAMP}. In this paper, we analyze the specific threat of an attacker that attempts to change the results of a U.S. election using adversarial examples printed on ballots. We make the following major contributions:
\begin{enumerate}
    \setlength{\itemsep}{0pt}
    \setlength{\parskip}{0pt}
    \setlength{\parsep}{0pt}
\item \textbf{Probabilistic Election Attack Framework} – We model the election process with ballot machine learning classifiers by developing a probabilistic framework. Our modeling encapsulates the uncertainty in both voter choice and the subset of which ballots are compromised. Using our framework, we derive a function for determining the proportion of compromised ballots needed to flip an election. This function depends only on the confidence bounds $z_{\alpha}$, number of ballots cast $N$ and underlying probabilities of voting for each candidate.
\item \textbf{Digital Adversarial Example Analyses} – We analyze digital adversarial example attacks on four different ballot classifiers using six $\epsilon$ values for each attack. We determine which type of ballot training dataset (marked/empty bubbles or including questionable marks) yields acceptable model performance. We empirically demonstrate that in the digital domain the $l_{2}$ and $l_{\infty}$ attacks are the most effective. We ascertain the perturbation parameters for imperceivable adversarial examples for each attack.
\item \textbf{Physical Adversarial Example Analyses} – We physically print and scan six different adversarial example attacks including $l_{\infty}$-APGD, $l_{2}$-APGD, $l_{1}$-APGD, $l_{0}$ PGD, $l_{0}+l_{\infty}$ PGD and $l_{0}+\sigma$-map PGD on four different models, CaiT, ResNet, VGG and SVM. We run these experiments to determine which attack is most effective when physically printed. Our analyses in the physical domain reveal that $l_{1}$ is the most effective (with $l_{2}$ also effective), in contrast to the results in the digital domain. This demonstrates that accurate evaluation of any future attack or defense design in the voting domain requires physical experimentation.
\end{enumerate}
\textbf{Disclaimer: } We attack commonly used machine learning models. We do not target any model that used in the U.S. election pipeline. The purpose of this work is to analyze and understand the risk in potentially using machine learning models for ballot classification. We seek to give election officials a more comprehensive understanding of emerging voting technologies and their potential security risks.

\textbf{Paper Organization.} In Section~\ref{sec:data}, we discuss the datasets and models analyzed in this work, as well as the related voting literature. A description of our adversarial threat model and the U.S. election pipeline is given in Section~\ref{sec:threatModel}. In Section~\ref{sec:derivation}, we construct a probabilistic framework for modeling the compromised and uncompromised election process. We derive a relationship between the number of compromised adversarial example ballots and the probability of successfully flipping an election. In Section~\ref{sec:digital}, we evaluate adversarial example attacks in the digital domain. In Section~\ref{sec:physical}, we analyze the empirical results from printing and scanning six different adversarial example attacks. Lastly, we offer concluding remarks in Section~\ref{sec:conclusion}.
\section{Datasets, Models and Related Voting Work}
\label{sec:data}
In this section we briefly discuss related election security work, the details related to the ballot datasets used in our analyses and the machine learning models we trained. All datasets and trained models will be released upon publication.

Using Convolutional Neural Networks (CNNs) as a replacement for optical scanners was first proposed in~\cite{CNNBallot}. For training machine learning models, there are three publicly available ballot image datasets. The \href{https://electionstransparencyproject.com/}{Humboldt County dataset} is unlabeled, and therefore not applicable to supervised learning. The \href{https://county.pueblo.org/clerk-and-recorder-department/ballot-images}{Pueblo County dataset} only allows access to individual images. To the best of our knowledge, no known framework exists to compile all images in this system. Lastly, the UConn Bubbles with Swatches dataset \cite{UConnVoterCenterv1_2025} is public, labeled and accessible. However, this dataset was created with Gaussian noise images. This was done to approximate the marginal marks (checks, crosses, pen rests) made by voters~\cite{Busting}. We use an updated dataset, UConn Bubbles with Marginal Marks dataset \cite{UConnVoterCenterv2_2026}, which contains synthetically generated marginal marks such as pen rests, checkmarks, crosses, straight scribbles, and random scribbles. Correctly recognizing actual images of marginal marks is the key task ballot classifiers are expected to complete in real-world applications~\cite{bajcsy2015systematic,CNNBallot,voter2022audit}. 

\textbf{Datasets}: We experiment with two ballot datasets from the UConn Bubbles with Marginal Marks dataset~\cite{UConnVoterCenterv2_2026}. Both datasets are comprised of $40 \times 50$ grayscale images. We denote our first dataset as the \textit{Combined} dataset. This dataset includes blank bubbles, voter-filled bubbles, and synthetically generated marginal marks. Marginal marks include pen rests (these are small marks from resting a pen that is not intended to be a mark), check marks, crosses, line-filled marks, and scribble marks. Marginal marks are hand labeled by the UConn Voter Center. Example marginal mark images and additional dataset details can be found in the appendix. We denote the second dataset as the \textit{Bubbles} dataset. This dataset includes only blank ballot bubbles and filled ballot bubbles, with no marginal marks. We include results corresponding to the Bubbles dataset to demonstrate that training on blank/filled bubbles alone does not yield high accuracy on marginal marks. The Combined dataset contains 42,679 training images and 10,652 validation images. The Bubbles dataset contains 28,589 training images and 7,137 validation images.

\textbf{Models}: For ballot classification, we train and attack four different models. We choose a selection of models that varies both the architecture and model complexity to have a clear understanding of model robustness in the voting domain. In order of model complexity, we start with a simple linear Support Vector Machine (SVM)~\cite{SVM,SVM2} with 2,001 trainable parameters. We include this model to explore the robustness of simple, low complexity models. Ballot recognition is an image classification problem, and CNNs are some of the most commonly used models for this type of task. We train two CNNs, a Very Deep Convolutional Network (VGG-16)~\cite{VGG16} with 14,723,010 trainable parameters and a ResNet-20 model~\cite{RESNET} with 568,033 trainable parameters. VGG and ResNet are some of the most commonly used CNN architectures for image classification. Vision Transformers have emerged as an alternative to CNNs for image classification~\cite{VIT}. We test one state-of-the-art Transformer, the Class-Attention in Image Transformers (CaiT)~\cite{CAIT}. Our CaiT model has 56,730,626 trainable parameters. Training details for each model can be found in the appendix.

\section{Adversarial Threat Model}
\label{sec:threatModel}
\begin{figure*}[t]
\includegraphics[width=1\textwidth]{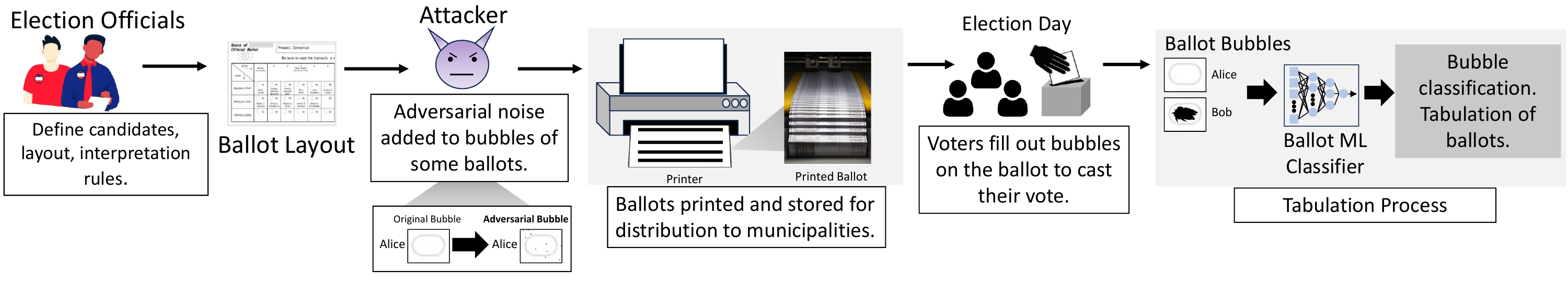} 
\caption{U.S. Election Pipeline: First officials define a ballot layout which is then sent for printing. During the printing step, an attacker could potentially compromise ballot printers and add adversarial examples to the ballots. The ballots are then distributed to voters on election day and tabulation can be done by a ballot classifier.}
\label{fig:overview}
\end{figure*}

We assume a threat model in which the attacker has two main capabilities. First, we assume the attacker has white-box access to the machine learning models used for ballot classification. This assumption is realistic and practical because knowledge of the machine learning model’s weights and architecture are necessary for openness and trust in the election process. Specifically, United States voting equipment standards~\href{https://www.eac.gov/sites/default/files/TestingCertification/Voluntary_Voting_System_Guidelines_Version_2_0.pdf}{VVSG 2.0} state, \textit{``Manufacturers must not make their specifications proprietary.''} The attacker's second power in our threat model is the ability to compromise one or more ballot printers, such that adversarial noise can be added to the bubbles of the ballots. Traditionally ballot printers are small vendors without robust cybersecurity teams in comparison to election equipment vendors, which are considered critical infrastructure. In terms of compromised printing, the attacker only adds visually imperceivable noise to the ballot bubbles. The magnitude of the adversarial noise is controlled such that a visual inspection of the ballot would not lead a voter to believe that the ballot had been filled in or tampered with. An overview of the election pipeline and the step in which a malicious attacker would act to insert adversarial noise is shown in Figure~\ref{fig:overview}. 

\textbf{Attack Setup:} In this work, experimentally evaluate both the under and over attacks first proposed in~\cite{Busting}. However, in terms of theory and attack formulation, we focus on the \textit{over} attack. In this setup, the attacker compromises one or more printers used to print ballots. The adversary can use the compromised printer to add adversarial noise to the bubbles on a certain number of ballots. In the \textit{over} attack formulation, noise is only added to blank ballots before they are sent out to voters. No ballots can be manipulated after they have been distributed to polling places. The goal of the attacker is to change the results of the election in favor of their candidate. The attack is considered successful if the attacker's candidate receives more votes than any other candidate. 
\subsection{White-Box Adversarial Evasion Attacks}
\label{subsec:whitebox}
The standard white-box adversarial example evasion attack~\cite{carlini2019evaluating} formulates the following optimization problem:
\begin{equation}
\label{eq:adv}
\max_{x_{adv}} L(x_{adv}, y; \theta)
\quad \text{s.t.} \quad
\|x - x_{adv}\|_p \le \epsilon
\end{equation}
where $L$ is the loss function, $x$ is the clean example with corresponding class label $y$ and $\theta$ is the variable representing the trained model parameters. The adversarial example is denoted as $x_{adv}=x+\delta$, where $\delta$ is the adversarial noise. In Equation~\ref{eq:adv}, the order of the norm is $p$ and $\epsilon$ is a parameter that determines the maximum amount of noise added to the clean example $x$. 

Maximizing the loss function using different $p$ and considering $\epsilon$ values produces adversarial examples that are visually distinct. In this work we evaluate the robustness of ballot classifiers with respect to six different norm attacks. We test the Auto Projected Gradient Descent (APGD) attack~\cite{APGD} with $p=2$ and $p=\infty$. We also run the sparse $l_{1}$-APGD attack~\cite{AttackL1}. For $p=0$, we test three different attacks proposed in~\cite{AttackL0}. These attacks are the $l_{0}$ Projected Gradient Descent (PGD) with $p=0$, $l_{0}+l_{\infty}$ PGD with double norm constraints ($p=0$ and $p=\infty$) and $l_{0}+\sigma$ map PGD ($p=0$).

Evaluating model robustness against attacks that use different $p$ values is essential for two reasons. First, it has been well established in the literature that only considering the most common norm attack $(p=\infty)$ does not guarantee robustness to other norm attacks~\cite{Union, AttackL0}. Second, the printing process for ballots is noisy, complex, and non-deterministic. It is an open question as to which norm attack is most effective when physically printed.

\section{Probabilistic Election Attack Framework}
\label{sec:derivation}
Two fundamental questions exist regarding the execution of attacks on elections with adversarial examples. First, how many ballots need to be printed with the most effective $l_{p}$ adversarial example for the election result to change? Second, what type of $l_{p}$ adversarial example attack is most effective? In this section, we answer the first question through analysis of the election as a probabilistic process. By explicitly modeling uncertainty in voter behavior and the number of compromised ballots, the resulting probabilities allow one to assess the likelihood of a successful attack (changing an election outcome). Specifically, we derive an equation that relates the confidence bounds for changing an election result with the number of compromised adversarial example ballots.

\subsection{The Election Process Framework}

Consider an election with $N$ voters and two candidates ($A$ and $B$). Let $X_1,\ldots,X_N$ denote the true (uncompromised) votes, where each vote is modeled as a discrete random variable $X_i$. With probability $p_{a}$, a voter casts a ballot for candidate~A and $X_i =1$. With probability $p_{b}$, a voter casts a ballot for candidate B and $X_i =-1$. The voter leaves the choice for this race blank with probability $p_{0}$, in which case $X_i =0$. Therefore, the vote probability distribution is specified by:
\[
p_a = \mathbb{P}(X_i = -1), 
p_b = \mathbb{P}(X_i = 1),
p_0 = \mathbb{P}(X_i = 0),
\]
with $p_a + p_b + p_0 = 1$.
The true vote margin is defined as $\Delta = p_b - p_a > 0$. This implies, without loss of generality, that candidate~$B$ is the true expected winner in the absence of compromised ballots (an attack with adversarial examples). Equivalently, the distribution can be parameterized by $(p_b, \Delta)$, with the remaining probabilities given by $p_a = p_b - \Delta$ and $p_0 = 1 - 2p_b + \Delta$. The expectation of the vote distribution under an uncompromised election process is \[\mathbb{E}[X_i] = p_b - p_a = \Delta.\] The variance of the vote distribution under an uncompromised election process is: \[\mathrm{Var}(X_i)=2p_b - \Delta - \Delta^2.\] Assuming that the $X_i$ variables are independent and identically distributed, the law of large numbers implies that as $N \to \infty$, the empirical difference in the proportion of votes $\bar{x} = \frac{1}{N} \sum_{i=1}^{N} X_i$ converges in probability to the true margin $\Delta$.

\subsection{The Compromised Election Process Framework}
The goal of the attacker is to compromise the election so that candidate $A$ wins. The attacker accomplishes this by manipulating a certain subset of the $N$ ballots cast by the voters. We define $Y_i$ as a Bernoulli random variable that indicates whether the ballot cast by the voter $i$ is compromised. Here compromised means that the ballot has been printed with adversarial noise added to the blank bubble associated with candidate $A$. $Y_i =1$, if ballot $i$ is compromised and $Y_i =0$ otherwise. Let $p_c = \mathbb{P}(Y_i=1)$ characterize the population fraction of the ballots that have been compromised. We denote $W_i$ as the observed vote associated with the ballot $i$ after a subset of the ballots is compromised. For voters who receive an uncompromised ballot, the observed vote always coincides with the true vote: $\{W_i \mid Y_i = 0 \} = X_i$. When a voter casts a compromised ballot, we have the following distribution:
\begin{equation}
\label{eq:comp}
\{ W_i \mid Y_i = 1 \} =
\begin{cases}
-1, & \text{with probability } 1-2p_b+\Delta, \\
-1, & \text{with probability $p_{b}-\Delta$} ,\\
0, & \text{with probability } p_{b}.
\end{cases}  
\end{equation}
There are three possible cases when an election system processes a compromised ballot as given by Equation~\ref{eq:comp}. When a voter leaves the race blank, the presence of an adversarial signal causes the ballot classifier to detect that the bubble for candidate~$A$ has been filled. A vote is counted for candidate~$A$ in the blank case and this event occurs with probability $ 1-2p_b+\Delta$. If a voter fills in the bubble for candidate~$A$, they overwrite the adversarial signal and a vote is still cast for candidate~$A$. The overwrite case occurs with probability $p_{b}-\Delta$. The final case is when the voter fills in the bubble for candidate~$B$ but the compromised ballot also contains an adversarial signal on the bubble for candidate~$A$. How election officials handle this case is municipally and situationally dependent~\cite{sinclair2003overvotes,neely2008whose}. For consistency, we assume the worst case scenario for the legitimate election process: the "overvote discard" outcome. Under this process, ballots that are improperly marked are discarded. This case occurs with probability $p_{b}$. We further discuss other possible derivations based on different election process assumptions in the appendix.  

\subsection{Attack Confidence Function}

Given the compromised election framework developed in the previous subsection, we derive the fraction of compromised ballots $p_{c}$ as a function of the probability of the attack succeeding $(1-\alpha)$. E.g., $(1-\alpha)=95\%$ means the attacker has a $95\%$ probability that candidate $A$ will win the election when $N\times p_{c}$ ballots are compromised using adversarial examples. 

By definition of the random variable $W_i$, each observed vote for candidate~$A$ contributes $-1$ to the aggregate, while each observed vote for candidate~$B$ contributes $+1$. Thus, the observed mean $\bar w = \frac{1}{N}\sum_{i=1}^N W_i$ is the difference between the proportion of votes obtained by candidate~$B$ and the proportion of votes obtained by candidate~$A$. Assuming that the true votes $X_i$ are independent across voters and that the compromise indicators $Y_i$ are independent, the observed votes $W_i$ are also independent. Under these assumptions, the Central Limit Theorem implies that for large $N$, we can approximate the distribution of $\bar w$ as:
\begin{equation}
 \bar w 
\;\sim\;
\mathcal{N}\!\left(\mathbb{E}[W_i],\,\frac{\mathrm{Var}(W_i)}{N}\right).   
\end{equation}
By the law of total expectation:
\begin{multline}
\label{eq:ew}
\mathbb{E}[W_i]=
\mathbb{E}\!\left[\mathbb{E}(W_i \mid Y_i)\right]
\\=(1-p_c)\Delta + p_c\,\mathbb{E}(W_i \mid Y_i=1).
\end{multline}
The last term $p_c\,\mathbb{E}(W_i \mid Y_i=1)$ in Equation~\ref{eq:ew} can be evaluated using Equation~\ref{eq:comp} with the overvote discard assumption. Therefore, $\mathbb{E}[W_i]=\Delta - p_c d$ where $d := 1+\Delta-p_b$. The variance of $W_{i}$ can be determined using the law of total variance:
\begin{multline}
\label{eq:var}
\mathrm{Var}(W_i)
=\mathbb{E}\!\left[\mathrm{Var}(W_i \mid Y_i)\right]+
\mathrm{Var}\!\left(\mathbb{E}[W_i \mid Y_i]\right)
\\=(1-p_c)v_0 + p_c v_1 + p_c(1-p_c)d^2,
\end{multline}
where the terms $v_0$ and $v_1$ are introduced for notational simplicity and are defined as $v_0 = 2p_b-\Delta-\Delta^2$ and $v_1 = p_b(1-p_b)$. When $\bar w<0$, this indicates that candidate~$A$ has received more votes than candidate~$B$ and that the attacker was successfully in changing the outcome of the election. We want to determine the probability of this event: $\mathbb{P}(\bar w<0)\ge 1 - \alpha$ in terms of $p_{c}$, the fraction of compromised ballots. Under the normal approximation this condition is equivalent to:
\begin{equation}
\label{eq:stepOne}
\mathbb{E}[W_i] + z_{\alpha}\sqrt{\frac{\mathrm{Var}(W_i)}{N}} \le 0,
\end{equation}
where $\mathbb{P} (Z > z_{\alpha}) = 1 - \alpha$ for any standard normal random variable $Z \sim N(0,1)$. Substituting $\mathbb{E}[W_i]=\Delta - p_c d$ and the computed variance from Equation~\ref{eq:var} into Equation~\ref{eq:stepOne} yields:
\begin{equation}
\label{eq:stepTwo}
\Delta - p_c d
+
z_{\alpha}
\sqrt{
\frac{
(1-p_c)v_0 + p_c v_1 + p_c(1-p_c)d^2
}{N}
}
\le 0.
\end{equation}
Solving Equation~\ref{eq:stepTwo} for $p_c$ gives the following closed form solution:
\begin{equation}
\label{eq:pcFinal}
p_c^\star
=
\frac{-k_1 + \sqrt{k_1^2 - 4k_2 k_0}}{2k_2},   
\end{equation}
where: \begin{align*}
k_2
&=
N d^2 + (z_{\alpha})^2 d^2,\\
k_1
&=
-2N\Delta d - (z_{\alpha})^2\bigl(v_1 - v_0 + d^2\bigr),\\
k_0
&=
N\Delta^2 - (z_{\alpha})^2 v_0.
\end{align*}
For brevity we omit the intermediate simplification steps and provide them in the appendix. 

\textbf{Derivation Significance:} Equation~\ref{eq:pcFinal} allows an attacker to determine the expected number of ballots ($N\times p_{c}^*$) to be printed with adversarial examples, such that the attacker wins with confidence $1-\alpha$. Previous work~\cite{Busting} indicated that a successful attack was plausible using data from past U.S. elections, but never created a generalized probabilistic framework. In our derivation, the most essential parameter in the attack, $p_{c}^*$, only depends on the probability of candidate~$B$ winning $p_{b}$, the uncompromised election difference $\Delta$, the number of ballots cast $N$ and the desired attacker confidence represented by the standard normal critical value $z_{\alpha}$. In U.S. elections these values can all be estimated from polling or past public election data.
\section{Digital Adversarial Attacks}
\label{sec:digital}
In this section, we analyze the clean accuracy and adversarial robustness of four models machine learning models in the digital domain. The digital domain is significant because it represents the best case for an attacker creating adversarial examples, as no printing noise is present. We train four models of varying complexity: SVM~\cite{SVM,SVM2}, VGG-16~\cite{VGG16}, ResNet-20~\cite{RESNET}, and CaiT~\cite{CAIT}. There are two possible ways to train each model based on which dataset is used. The Bubbles dataset contains only filled and blank bubble images. We denote models trained on the Bubbles dataset with the suffix ``-B'' (e.g., SVM-B). The combined dataset contains blank and filled bubbles, as well as the more challenging marginal marks (see Section~\ref{sec:data} for full dataset details). We denote a model trained on the Combined dataset with the suffix ``-C'' (e.g., SVM-C). Full model training details and related hyperparameters are given in the appendix.  

\begin{table}[h]
\small
\begin{centering}
\resizebox{\columnwidth}{!}{%
\begin{tabular}{l|rr|rr|}
\hline
\multirow{2}{*}{Model (Trained On)} & \multicolumn{2}{c|}{Training Accuracy} & \multicolumn{2}{c|}{Validation Accuracy} \\ \cline{2-5}
& Bubbles & Combined & Bubbles & Combined \\ \hline
SVM-B        & 1.0000 & .8992 & 1.0000 & .9026 \\
VGG-16-B      & 1.0000 & .9205 & 1.0000 & .9247 \\
ResNet-20-B  & 1.0000 & .9147 & 1.0000 & .9155 \\
CaiT-B      & 1.0000 & .8955 & 1.0000 & .8992 \\ \hline
SVM-C        & .9989 & .9824 & .9989 & .9784 \\
VGG-16-C      & 1.0000 & .9995 & 1.0000 & .9972  \\ 
ResNet-20-C  & 1.0000 & 1.0000 & 1.0000 & .9972 \\
CaiT-C      & 1.0000 & 1.0000 & 1.0000 & .9961 \\ \hline
\end{tabular}}
\end{centering}
\caption{Clean training and validation accuracies on the Bubble and Combined dataset for all models.}
\label{table:CleanAcc}
\end{table}
\textbf{Clean Evaluation:} The clean training and validation accuracy for each model is given in Table~\ref{table:CleanAcc}. There are three important takeaways from this set of experiments. First, all the “-B” models are able to achieve $100\%$ accuracy on the Bubbles validation set. \textit{Second, training on only the Bubbles dataset does not yield good performance on the Combined validation dataset}. No “-B” model achieves $>93\%$ on the Combined validation data. This is significant because it illustrates an important utility concept. Training on blank and filled bubbles alone is not sufficient for real world applications, where voters may cast ballots with marginal marks. The third important result from this set of experiments is that training on the Combined dataset yields high accuracy on both the Combined and Bubbles validation sets. For example, all “-C” models are able to achieve $>99\%$ on the Bubbles validation dataset. CaiT-C, ResNet-C and VGG-16-C all achieve $>99\%$ on the Combined validation dataset. These results are also significant because the Combined dataset is the first labeled ballot dataset to include voter marginal marks (pen rests, check marks, crosses, line filled and scribble images). \textit{The high accuracy in the clean digital context demonstrates that machine learning models can be used to accurately assess voter intent on ballots, even when voters use marginal marks}.
\textbf{Experimental Attack Setup:} We evaluate the robustness of four models each trained on two different datasets: SVM-B, VGG-16-B, ResNet-20-B, CaiT-B, SVM-C, VGG-16-C, ResNet-20-C and CaiT-C. Each model is tested with six different adversarial attacks. Each attack is run with six different noise values (perturbation parameters). The purpose of this set of experiments is two-fold. First, ballot classification using machine learning is a newly developing field that requires standards. For example, the CIFAR-10 dataset has a well established standard imperceivable attack $(\epsilon=8/255)$ parameter value in the literature~\cite{APGD,mahmood2021robustness}. Standards are not yet well established for the ballot datasets for each of the different norm attacks. It is important to experiment with a variety of noise budgets to determine a standard, imperceivable noise values. The second important result of these experiments is to assess which of the six attacks performs best in the digital domain. This is significant because a real attacker is likely to utilize the most effective adversarial example attack.

\textbf{Attack Parameters:} The six attacks we test are \(l_{\infty}\)-APGD, \(l_{1}\)-APGD, \(l_{2}\)-APGD, \(l_{0}\) PGD, \(l_{0}+l_{\infty}\) PGD, and \(l_{0}+\sigma\)-map PGD. For each attack, we test six different noise budget values. Specifically, for \(l_{\infty}\)-APGD we use \(\epsilon \in \{4,8,16,32,64,255\}/255\); for \(l_{1}\)-APGD 
 we use \(\epsilon \in \{2,4,8,20,200,2000\}\); for \(l_{2}\)-APGD we use \(\epsilon \in \{1,2,3,5,15,45\}\). For the sparse attacks \(l_{0}\) PGD and \(l_{0}+\sigma\)-map PGD we use \(k \in \{1,10,20,200,500,2000\}\), where \(k\) is the maximum number of perturbed pixels. For \(l_{0}+l_{\infty}\) PGD we fix $k=20$ and use \(\epsilon \in \{4,8,16,32,64,255\}/255\).
 
 All remaining attack hyperparameters (number of steps, step size, etc.) are fixed per attack and are provided in the appendix. Adversarial examples for each attack are generated by starting from 1000 \emph{clean} (i.e., unperturbed) images that are class-wise balanced. In all digital experiments, the starting clean images come from the Bubbles validation set only. This matches prior voting-domain evaluation practices~\cite{Busting}.

\begin{figure*}
\begin{center}
\includegraphics[width=0.6\textwidth]{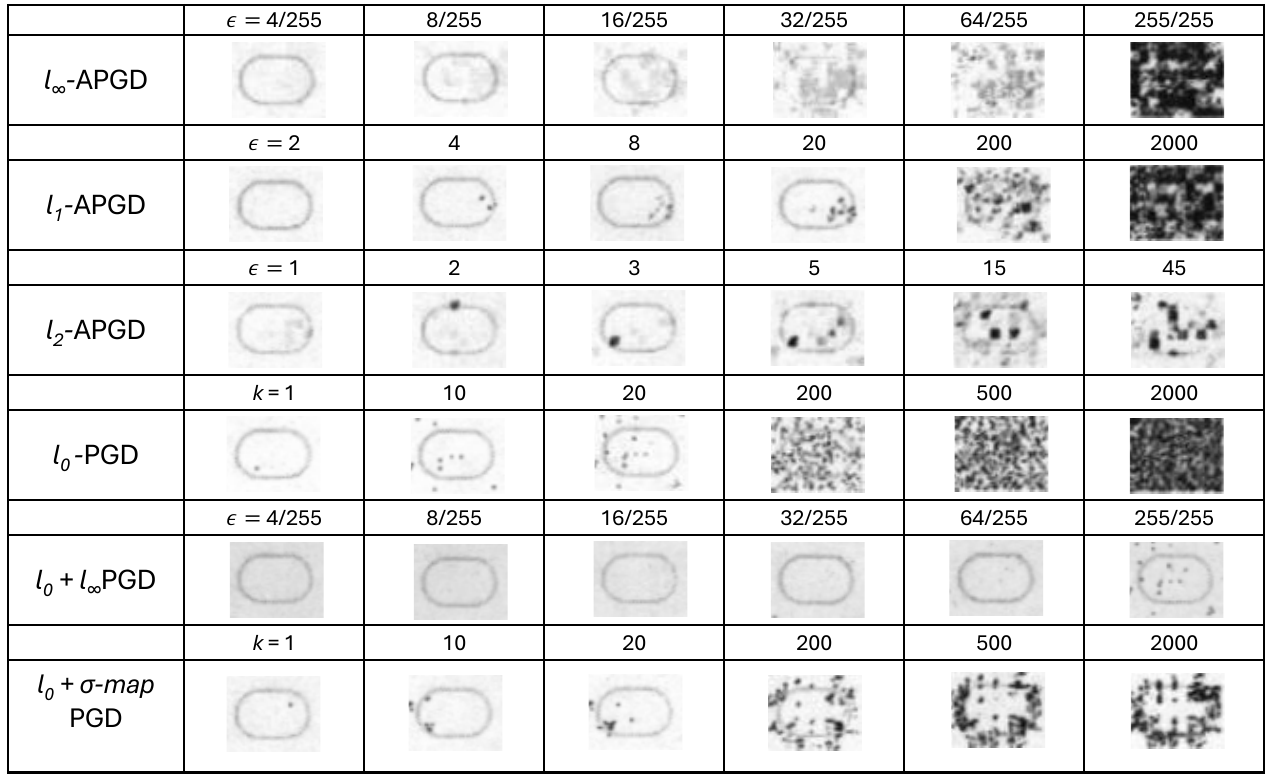} \end{center} \caption{Physical Adversarial examples with varying $\epsilon$ or $k$ for all attacks on the CaiT-C model.} \label{fig:varyModelFig} \end{figure*}

\begin{table*}[t]
\centering
\small
\setlength{\tabcolsep}{4pt}
\begin{tabular}{lcccccc}
\toprule
\textbf{Model} &
\textbf{$l_{\infty}$-APGD} &
\textbf{$l_{1}$-APGD} &
\textbf{$l_{2}$-APGD} &
\textbf{$l_{0}$ PGD} &
\textbf{$l_{0}{+}l_{\infty}$ PGD} &
\textbf{$l_{0}{+}\sigma$ PGD} \\
& {\footnotesize $\epsilon{=}16/255$} &
{\footnotesize $\epsilon{=}20$} &
{\footnotesize $\epsilon{=}2$} &
{\footnotesize $k{=}1$} &
{\footnotesize $\epsilon{=}64/255$} &
{\footnotesize $k{=}1$} \\
\midrule
SVM-B      & 1.000 & 1.000 & 1.000 & 1.000 & 1.000 & 1.000 \\
SVM-C      & \textbf{0.500} & \textbf{0.500} & \textbf{0.500} & 1.000 & 1.000 & 1.000 \\  \hline

VGG16-B    & 0.000 & 0.498 & 0.000 & 1.000 & 1.000 & 0.774 \\
VGG16-C    & \textbf{0.000} & 0.500 & \textbf{0.000} & 0.889 & 0.537 & 0.500 \\  \hline
ResNet20-B & 0.219 & 0.027 & 0.260 & 1.000 & 1.000 & 1.000 \\
ResNet20-C & \textbf{0.000} & \textbf{0.000} & \textbf{0.000} & 0.998 & 0.500 & 0.558 \\  \hline
CaiT-B     & 1.000 & 0.496 & 0.999 & 1.000 & 1.000 & 1.000 \\
CaiT-C     & 0.500 & 0.499 & \textbf{0.498} & 1.000 & 0.997 & 0.930 \\
\bottomrule
\end{tabular}
\caption{Robust accuracy for each attack using the selected imperceptible perturbation parameters. For the "-C" models, we bold the attack that is the most effective (lowest robustness).}
\label{tab:robust-imperceptible}
\end{table*}

\textbf{Perturbation Parameters for Imperceivable:} 
For each of the different norm attacks, 
we generate adversarial examples using a range of different perturbation parameters ($\epsilon$ and $k$). 
Figure~\ref{fig:varyModelFig} illustrates how the visibility of the adversarial noise varies across different perturbation parameters for each attack. 
The general trend is that as the magnitude of the perturbation parameter increases, the more noticeable the added noise. 
The visual appearance of the noise also depends on the norm constraint. In Figure~\ref{fig:varyModelFig}, $l_{\infty}$ and $l_{2}$ attacks produce 
square noise patterns that become increasingly prominent as $\epsilon$ grows. $l_{1}$ noise remains relatively localized at small $\epsilon$ but 
becomes more noticeable at larger budgets. In contrast, $l_{0}$-based attacks modify only a subset of pixels controlled by $k$. 
Based on visual appearance of attacks, we chose imperceivable levels of noise for each attack using majority voting from a panel of 6 U.S. 
election security experts who reviewed physical adversarial examples across all attacks on all 4 Combined models. 
The imperceptible values used in this paper are:
\(l_{\infty}\)-APGD \(\epsilon=16/255\),
\(l_{1}\)-APGD \(\epsilon=20\),
\(l_{2}\)-APGD \(\epsilon=2\),
\(l_{0}\) PGD \(k=1\),
\(l_{0}+l_{\infty}\) PGD \(\epsilon=64/255\),
and \(l_{0}+\sigma\)-map PGD \(k=1\).
Additional details regarding the robustness of each model for all perturbations values is given in the appendix.
\subsection{Digital Experimental Analyses}
In Table~\ref{tab:robust-imperceptible} we report the robustness of each model to the six different norm attacks. Each attack uses the imperceivable perturbation parameters established earlier in this section.
For all the “-C” models, the most effective attack in the digital domain was the $l_{2}$ attack. 
This attack performed best on CaiT-C with $0.498$ robustness. It achieved $0.0$ robustness on both VGG16-C and ResNet20-C (with $l_{\infty}$ also achieving $0.0$ on both models). For SVM-C, the $l_{\infty}$, $l_{1}$, and $l_{2}$ attacks performed equally well, each yielding $0.500$ robustness. The significance of these results is that in the digital domain for ballot classification, the most effective norm attack is \textit{ dependent on the choice of model.} 
For the \(l_{0}\)-based attacks, our results indicate they are often less effective than $l_{1}$, $l_{2}$, and $l_{\infty}$. 
For example, $l_{0}$ PGD yields $1.000$ on SVM-C and $1.000$ on CaiT-C, suggesting that larger perturbation budgets (larger $k$) 
are needed to reliably reduce the robustness of these models. 
It is common for  models to misclassify all \emph{Over} examples (switching non-marks to marks) at an $\epsilon$ level and correctly classify all \emph{Under} examples (switching marks to non-marks).

\section{Physical World Attacks}
\label{sec:physical}
In this section we analyze the robustness of the models trained on the Combined dataset to adversarial examples that are physical realized through printing and scanning. We narrow our focus to the ``-C'' models. This is due to the fact that only the ``-C'' models have satisfactory performance on bubbles and marginal voter marks, which would be encountered in a real election.
\begin{table*}[t]
\centering
\small
\setlength{\tabcolsep}{3pt}
\begin{tabular}{lcccccccc}
\toprule
\textbf{Model} &
\textbf{Clean} &
\textbf{Clean+AE} &
\textbf{$l_{\infty}$-APGD} &
\textbf{$l_{1}$-APGD} &
\textbf{$l_{2}$-APGD} &
\textbf{$l_{0}$ PGD} &
\textbf{$l_{0}{+}l_{\infty}$ PGD} &
\textbf{$l_{0}{+}\sigma$ PGD} \\
& {\footnotesize $0/255$} &
{\footnotesize $0/255$} &
{\footnotesize $\epsilon{=}16/255$} &
{\footnotesize $\epsilon{=}20$} &
{\footnotesize $\epsilon{=}2$} &
{\footnotesize $k{=}1$} &
{\footnotesize $\epsilon{=}64/255$} &
{\footnotesize $k{=}1$} \\
\midrule
SVM-C      & 0.936 & 0.977 & 0.583 & 0.603 & \textbf{0.552} & 1.000 & 0.960 & 1.000 \\
VGG-16-C    & 0.971 & 0.972 & 1.000 & \textbf{0.708} & 0.946 & 1.000 & 1.000 & 0.999 \\
ResNet20-C & 0.972 & 0.976 & 1.000 & \textbf{0.971} & 0.989 & 1.000 & 1.000 & 0.975 \\
CaiT-C     & 0.969 & 0.994 & 0.540 & \textbf{0.501} & 0.503 & 1.000 & 0.919 & 0.963 \\
\bottomrule
\end{tabular}
\caption{Physical world robust accuracy for the Combined models for each of the six norm attacks.}
\label{tab:robust-physical}
\end{table*}

\begin{figure}
\centering
\includegraphics[width=0.48\textwidth]{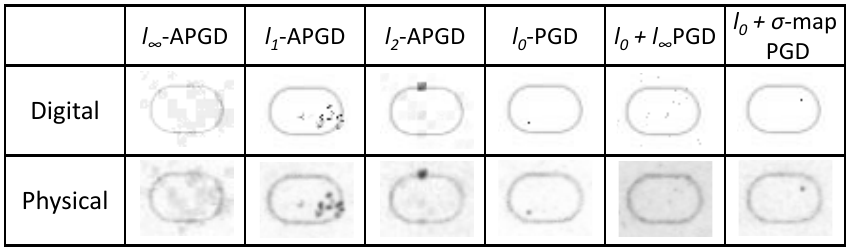}
\caption{Digital and printed physical adversarial examples for the CaiT-C model. Each attack uses the imperceivable parameters determined in Section~\ref{sec:digital}.}
\label{fig:virtual_physical_caitc}
\end{figure}

\textbf{Printing Pipeline.} The printing process begins with the selection of $1000$ correctly identified validation examples from the Bubbles dataset ($500$ blank and $500$ filled bubbles). The bubbles are shuffled to allow a mixture of empty and filled bubbles to be printed on each page. This process is done to avoid contrast enhancement issues that occur when printing and scanning only empty bubbles on a page. In addition, a black bar is printed at the bottom of the page to allow the scanner to calibrate brightness contrast levels. For printing we lay out each bubble image on an $8.5 \times 11$ inch letter-sized paper, with a maximum of 96 bubble images per page. Each sheet is printed using a HP LaserJet Pro MFP 3101fdw (laser printer). After printing, each sheet is scanned using a Ricoh fujitsu fi-7160 high-fidelity scanner and a corresponding .TIFF file is generated. Alignment of the images is accomplished with Enhanced Correlation Coefficient (ECC) alignment and an affine transformation implemented in OpenCV~\cite{bradski2000opencv}. Segmentation is performed using the geometry of the printed bubbles on the sheet before printing, i.e. the page height and width, and the bubble image height and width. This segmentation on aligned sheets extracts the original $40\times 50$ bubbles which are then converted back to a PyTorch tensor for further processing. To enhance the clean accuracy of each model and reduce printing noise, physically scanned images are passed through a denoising pipeline. Our denoising pipeline is made up of a U-Net autoencoder~\cite{ronneberger2015u} that is trained with digital training data and training data that has been printed and scanned. The accuracy pre-denoising pipeline and post-denoising pipeline is given in the second and third columns of Table~\ref{tab:robust-physical}. In Figure~\ref{fig:virtual_physical_caitc} we show examples of digital adversarial examples for each attack and their printed and scanned counterparts. 

\subsection{Physical Experimental Analyses}
In Table~\ref{tab:robust-physical},we report the robustness of SVM-C, VGG-16-C, ResNet-20-C and CaiT-C for each of the six different norm attacks. For each attack we use the imperceptible perturbation values that were established in Section~\ref{sec:digital}. In the appendix, we provide robustness measurements for every attack at every perturbation value (in total 144,000 adversarial examples printed and evaluated). There are three key results from the physical world experiments. First, in the physical domain for VGG-16-C, ResNet-20-C, CaiT-C, the $l_{1}$-APGD attack was the most effective (lowest robustness). For SVM-C, the most effective attack was $l_{2}$-APGD with a robustness of $55.2\%$. Overall, CaiT was the least robust of all models (considering the minimum robustness measurement across all attacks). This second finding is significant because it indicates that in the voting domain, increasing model complexity alone does not offer additional security. A third trend is apparent from the printing and scanning of adversarial examples: There is a difference between the most effective adversarial examples in the digital domain and the physical domain. 
In the physical domain, the $l_{1}$-APGD attack is most effective for three of the four models, with $l_{2}$-APGD most effective for SVM-C, whereas in the digital domain, the $l_{2}$ and $l_{\infty}$ norm attacks were most effective. Our findings highlight a key point, \textit{future adversarial machine learning research in the voting domain must include physical adversarial example realizations.} Experimenting in the digital domain alone does not encapsulate the complexity and significant effects of the printing process. Why are certain attacks more effective in the physical domain? In Table~\ref{table:rmse} we computed the difference between the digital and printed adversarial noise $\delta$ for each norm attack. For each attack we averaged this difference for the 1000 printed and digital examples and further averaged the results over all 4 models. We tested several different metrics, including Root Mean Square Error (RMSE), Kullback–Leibler Divergence (KL) and Structural Similarity Index Measure (SSIM)~\cite{SSIM}.
\begin{table}[h]
\tiny
\centering
\begin{tabular}{r|c|c|c|c|c|c|}
\cline{2-7}
\multicolumn{1}{l|}{} &
  \multicolumn{1}{l|}{$l_{\infty}$} &
  \multicolumn{1}{l|}{$l_{1}$} &
  \multicolumn{1}{l|}{$l_{2}$} &
  \multicolumn{1}{l|}{$l_{0}$} &
  \multicolumn{1}{l|}{$l_{0}+l_{\infty}$} &
  \multicolumn{1}{l|}{$l_{0}+\sigma$-map} \\ \hline
\multicolumn{1}{|r|}{RMSE} &
  \multicolumn{1}{c|}{0.062} &
  \multicolumn{1}{c|}{0.069} &
  \multicolumn{1}{c|}{0.058} &
  \multicolumn{1}{c|}{\textbf{0.054}} &
  \multicolumn{1}{c|}{0.078} &
  \multicolumn{1}{c|}{0.068} \\ \hline
\multicolumn{1}{|r|}{KL}   & 5.71  & 12.18   & \textbf{0.52} & 22.70 & 22.49 & 22.71 \\ \hline
\multicolumn{1}{|r|}{SSIM} & 0.442 & 0.415 & 0.474   & \textbf{0.481} & 0.250 & 0.455    \\ \hline
\end{tabular}
\caption{Root Mean Square Error (RMSE), Kullback–Leibler divergence (KL) and Structural Similarity Index Measure (SSIM) between the printed perturbation $\delta_{print}$ and the digital perturbation $\delta_{digital}$ for each norm attack.}
\label{table:rmse}
\end{table}

\begin{figure}
\centering
\includegraphics[width=0.48\textwidth]{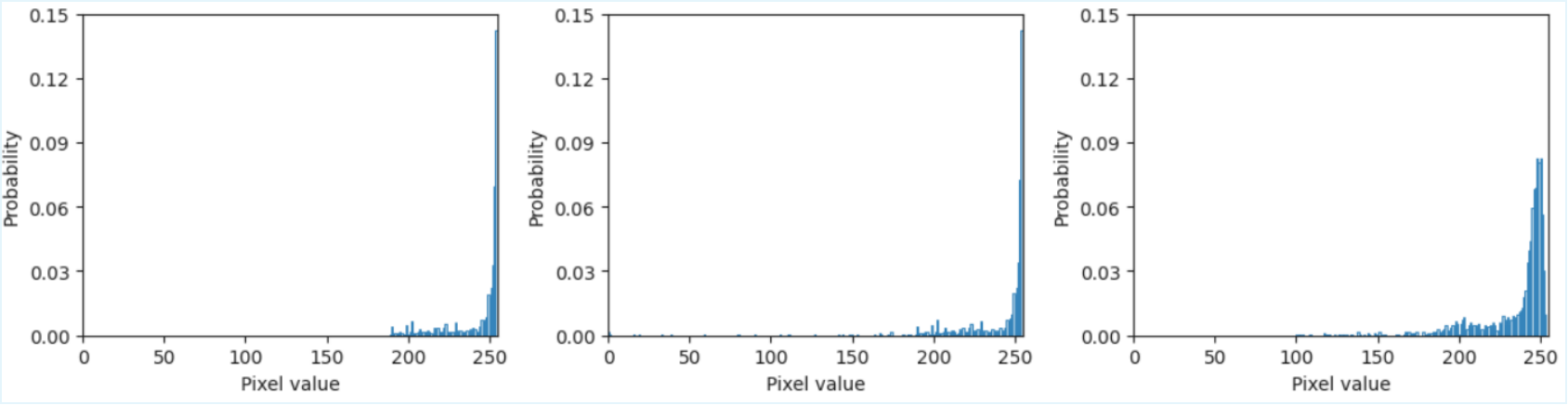}
\caption{Histograms for a digital clean ballot (left), its corresponding $l_{1}$ digital adversarial example (center) and the physically printed and scanned adversarial example (right).}
\label{fig:histogram}
\end{figure}

Overall the results from Table~\ref{table:rmse} reveal an interesting trend: Adversarial attacks that closely recreate the digital noise signal in the physical domain do not correlate with more successful attacks. The $l_{0}$ attack has the lowest RMSE and highest SSIM, and the $l_{2}$ attack has the lowest KL value. However, as our experiments demonstrated, $l_{1}$-APGD and $l_{2}$-APGD performed best in the physical domain while $l_{0}$-based attacks are ineffective at the imperceivable budget used here ($k{=}1$). This discrepancy may be in part due to the noisy and non-deterministic effects of printing, and because traditional signal metrics do not capture which perturbations remain decision-effective after physical realization. In Figure~\ref{fig:histogram}, we illustrate this by showing the change in histograms (pixel distributions) between one digital and physical ballot. To summarize, our analysis show that thus far, none of the three traditional signal metrics strongly correlate with attack effectiveness in the physical domain.
\section{Conclusion}
\label{sec:conclusion}
 In this work, we explore two critical issues related to adversarial example attacks in the voting domain. First, through the development of a probabilistic framework, we determine the proportion of compromised adversarial example ballots needed for an attacker to successfully change the outcome of an election. Second, we investigate which adversarial example attack is most effective when physically realized. We evaluate six different norm attacks including $l_{\infty}$-APGD, $l_{2}$-APGD, $l_{1}$-APGD, $l_{0}$ PGD, $l_{0}+l_{\infty}$ PGD and $l_{0}+\sigma$-map PGD on four different models (144,000 printed adversarial examples evaluated). We show that in the physical domain, the $l_{1}$ and $l_{2}$ adversarial examples are most effective. In summary, our contributions push the field of machine learning and voting security forward by theoretically and empirically demonstrating when and how adversarial examples could change election results.

\bibliography{example_paper}
\bibliographystyle{icml2026}

\newpage
\appendix
\onecolumn
\section{Statistical Derivations}

\paragraph{Central moments of true vote $X_i$'s distribution.}
The true vote $X_i$ is constructed to have the following distribution:
\begin{equation}
X_i =
\begin{cases}
-1, & \text{with probability } p_b - \Delta, \\
\phantom{-}1, & \text{with probability } p_b, \\
\phantom{-}0, & \text{with probability } 1 + \Delta - 2p_b .
\end{cases}
\end{equation}
The first central moment (expectation) of the true vote distribution is
\begin{equation}
\mathbb{E}[X_i]
= -1 \cdot (p_b-\Delta) + 1 \cdot p_b + 0 \cdot (1+\Delta-2p_b)
= \Delta .
\end{equation}

The second central moment (variance) of the true vote distribution is
\begin{align}
\mathrm{Var}(X_i)
&= \mathbb{E}[X_i^2] - \big(\mathbb{E}[X_i]\big)^2 \nonumber \\
&= (-1)^2(p_b-\Delta) + 1^2(p_b) + 0^2(1+\Delta-2p_b) - \Delta^2 \nonumber \\
&= 2p_b - \Delta - \Delta^2 .
\end{align}

\paragraph{Compromise indicator $Y_i$.}
Let $Y_i$ be a Bernoulli random variable indicating whether ballot $i$
is compromised:
\begin{equation}
Y_i =
\begin{cases}
1, & \text{with probability } p_c, \\
0, & \text{with probability } 1-p_c .
\end{cases}
\end{equation}

While the distributions of $X_i$ and $Y_i$ capture the randomness of the
true votes and the prevalence of compromised ballots, respectively, the
distribution of the observed vote $W_i$ depends on how the election
system processes ballots containing multiple marked choices. Such
ballots may arise from a legitimate vote for candidate~B combined with
adversarial manipulation. Different handling rules induce different
observation models.

\paragraph{Ballot discard assumption.}
Under the ballot discard assumption, votes containing multiple marked
choices in the relevant race are classified as overvotes and discarded.
Consequently,
\begin{equation}
\{W_i \mid Y_i=0\} = X_i ,
\end{equation}
while for compromised ballots,
\begin{equation}
\{W_i \mid Y_i=1\} =
\begin{cases}
0, & \text{with probability } p_b, \\
-1, & \text{with probability } 1-p_b .
\end{cases}
\end{equation}

\paragraph{Central moments of the observed vote $W_i$.}
The central moments of $W_i$ follow from the law of total expectation,
\begin{equation}
\mathbb{E}[W_i] = \mathbb{E}\!\left[\mathbb{E}[W_i \mid Y_i]\right],
\end{equation}
and the law of total variance,
\begin{equation}
\mathrm{Var}(W_i)
=
\mathrm{Var}\!\left(\mathbb{E}[W_i \mid Y_i]\right)
+
\mathbb{E}\!\left[\mathrm{Var}(W_i \mid Y_i)\right].
\end{equation}

The expectation of the observed vote distribution is
\begin{align}
\mathbb{E}[W_i]
&= (1-p_c)\,\mathbb{E}[X_i] + p_c\,\mathbb{E}[W_i \mid Y_i=1] \nonumber \\
&= (1-p_c)\Delta + p_c\bigl[-(1-p_b)\bigr] \nonumber \\
&= \Delta - p_c(\Delta + 1 - p_b) \nonumber \\
&= \Delta - p_c d , \; \text{where} \; d :=  1 + \Delta - p_b .
\end{align}

The variance of the conditional expectation is
\begin{align}
\mathrm{Var}\!\left(\mathbb{E}[W_i \mid Y_i]\right)
&= (1-p_c)(p_c d)^2 + p_c\big((1-p_c)d\big)^2 \nonumber \\
&= p_c(1-p_c)d^2 .
\end{align}

Let
\begin{equation}
v_0 := \mathrm{Var}(W_i \mid Y_i=0),
\qquad
v_1 := \mathrm{Var}(W_i \mid Y_i=1).
\end{equation}
Then
\begin{align}
v_0 &= \mathrm{Var}(X_i) = 2p_b - \Delta - \Delta^2, \\
v_1 &= p_b(1-p_b).
\end{align}
By conditioning on $Y_i$,
\begin{equation}
\mathbb{E}\!\left[\mathrm{Var}(W_i \mid Y_i)\right]
= (1-p_c)v_0 + p_c v_1 .
\end{equation}

Combining terms,
\begin{equation}
\mathrm{Var}(W_i)
= p_c(1-p_c)d^2 + (1-p_c)v_0 + p_c v_1 .
\end{equation}

\paragraph{Normal approximation for the sample mean.}
Recall that each observed vote $W_i$ takes values in $\{-1,0,1\}$,
representing a vote for candidate~A, a discarded (or blank) vote, and a
vote for candidate~B, respectively. Consequently, the sum
\[
\sum_{i=1}^N W_i
\]
is equal to the difference between the total number of observed votes for
candidate~B and the total number of observed votes for candidate~A.
Let $M_b$ and $M_a$ denote the number of observed votes cast for
candidates~B and~A, respectively. Then
\begin{equation}
\sum_{i=1}^N W_i = M_b - M_a .
\end{equation}
Dividing by the total number of votes $N$ yields
\begin{equation}
\bar w := \frac{1}{N}\sum_{i=1}^N W_i
= \frac{M_b - M_a}{N},
\end{equation}
which represents the difference between the proportions of observed votes for candidate~B and candidate~A.

Since the true votes $\{X_i\}_{i=1}^N$ are assumed to be independent 
across voters, and the compromise indicators $\{Y_i\}_{i=1}^N$ are also
independent, the observed votes $\{W_i\}_{i=1}^N$ are independent. 
Moreover, the random variables $W_i$ are identically distributed. 
Consequently, the Central Limit Theorem implies that, for sufficiently 
large $N$, the sample mean
$\bar w$ is approximately normally distributed with mean
$\mathbb{E}[W_i]$ and variance $\mathrm{Var}(W_i)/N$, that is,
\begin{equation}
\bar w
\;\dot\sim\;
\mathcal{N}\!\left(
\mathbb{E}[W_i],
\frac{\mathrm{Var}(W_i)}{N}
\right).
\end{equation}

\paragraph{Probability bound and quadratic condition.}
Let $z_\alpha$ be a standard normal critical value satisfying
\[
\mathbb{P}(Z > z_\alpha) = 1-\alpha,
\qquad Z \sim \mathcal{N}(0,1).
\]
Then condition $\mathbb{P}(\bar w < 0) \ge 1-\alpha$ is equivalent to
\begin{equation}
\mathbb{E}[W_i]
+
z_\alpha \sqrt{\frac{\mathrm{Var}(W_i)}{N}}
\le 0 .
\end{equation}
This helps to determine a generalized solution for $p_c$ that provides a 
high probability of flipping an election. For $1 - \alpha = 0.95$, the probability condition is equivalent to a 95\% probability of having candidate~A win the election.

Substituting the expressions above yields
\begin{equation}
\Delta - p_c d
+
z_\alpha
\sqrt{
\frac{
(1-p_c)v_0 + p_c v_1 + p_c(1-p_c)d^2
}{N}
}
\le 0 .
\end{equation}

Rearranging and squaring both sides gives
\begin{equation}
(p_c d - \Delta)^2
\ge
\frac{(z_\alpha)^2}{N}
\left[
(1-p_c)v_0 + p_c v_1 + p_c(1-p_c)d^2
\right].
\end{equation}

Multiplying by $N$ and expanding yields a quadratic inequality
\begin{equation}
k_2 p_c^2 + k_1 p_c + k_0 \ge 0,
\end{equation}
with coefficients
\begin{align}
k_2 &= Nd^2 + (z_\alpha)^2 d^2, \nonumber \\
k_1 &= -2N\Delta d - (z_\alpha)^2\bigl(v_1 - v_0 + d^2\bigr), \\
k_0 &= N\Delta^2 - (z_\alpha)^2 v_0, \nonumber 
\end{align}
where 
\[
d := 1 + \Delta - p_b, \quad v_0 := 2p_b - \Delta - \Delta^2, \quad \textrm{and} \quad 
v_1 := p_b (1 - p_b)
\]

The smaller positive root of this quadratic is
\begin{equation}
p_c^\star
=
\frac{-k_1 + \sqrt{k_1^2 - 4k_2 k_0}}{2k_2}.
\end{equation}
\section{Datasets and Model Training}

\begin{figure*}[h!]
\begin{center}
\includegraphics[width=0.85\textwidth]{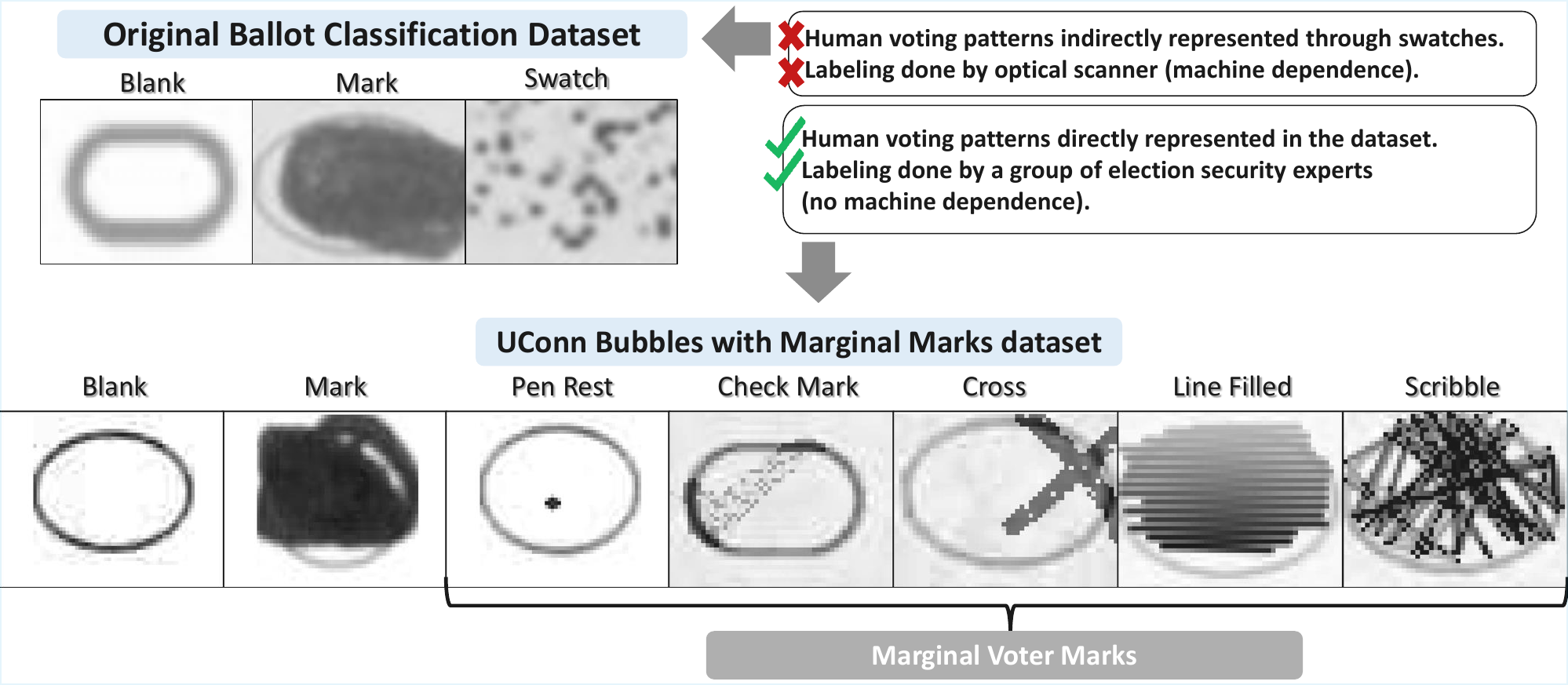} \end{center} \caption{UConn Bubbles with Marginal Marks dataset.} \label{fig:dataset} \end{figure*}

\paragraph{UConn Bubbles with Marginal Marks dataset:} Example images from the Combined dataset are shown in Figure~\ref{fig:dataset}. The dataset is publicly available~\href{https://zenodo.org/records/18344123}{here}.

\paragraph{SVM Training Details:} We trained linear Support Vector Machines (SVMs) on the OnlyBubbles and Combined datasets. 
Each input image of size $1 \times 40 \times 50$ was flattened into a 2{,}000-dimensional feature vector. 
Before training, we applied z-score normalization using a standard scaler (zero mean and unit variance). 
We trained SVMs using scikit-learn's \emph{SVC} with a \emph{linear} kernel and \texttt{class\_weight="balanced"} to address class imbalance. 
We selected the regularization parameter $C$ via a grid search over values ranging from $0.001$ to $10$, choosing the model with the best validation accuracy. 
To support gradient-based adversarial attacks, we ported the trained scikit-learn SVM into PyTorch. 
We extracted the learned coefficients and intercept from the model trained in standardized space and converted them back to the original pixel space. 
We then loaded these parameters into a PyTorch linear layer, and represented the binary classifier using two symmetric logits to ensure compatibility with attacks (e.g., APGD).

\paragraph{ResNet-20 training Details:} We train ResNet-20 models on both the Bubbles and Combined datasets. 
ResNet-20 is a standard residual CNN with 20 layers adapted to \(40\times 50\) single-channel inputs and two output classes.
Models are trained using cross-entropy loss with SGD. 
For the Bubbles dataset, we use a WeightedRandomSampler to address class imbalance during training.

\paragraph{CaiT Training Details:} We train CaiT models from on both the Bubbles and Combined datasets. 
We use a patch size of 5, embedding dimension of 512, 16 transformer layers (with 2 class-attention layers), 8 attention heads, and MLP dimension of 2048. 
Dropout and embedding dropout are set to 0.1, and layer dropout is 0.05. 
Models are trained with cross-entropy loss and the Adam optimizer. 
For the Bubbles dataset, we use a WeightedRandomSampler to address class imbalance.

\begin{table}[h]
\centering
\small
\begin{tabular}{l|ccccc}
\hline
\multirow{2}{*}{Model} & \multicolumn{5}{c|}{Hyperparameters} \\ \cline{2-6} 
 & LR & DR & Epochs & WD & BS \\ \hline
CaiT-B & 0.00005 & 0.1 & 200 & 0.0 & 128 \\
ResNet20-B & 0.1 & 0.0 & 60 & 0.0001 & 64 \\
VGG16-B & 0.001  & 0.0 & 40 & 0.0 & 64  \\ \hline
CaiT-C & 0.00005 & 0.1 & 200 & 0.0 & 128 \\
ResNet20-C & 0.1 & 0.0 & 100 & 0.0001 & 128 \\
VGG16-C & 0.001  & 0.0 & 40 & 0.0 & 64 \\ \hline
\end{tabular}
\caption{Hyperparameter settings for the various grayscale models. Abbreviations: LR = learning rate, DR = dropout rate, WD = weight decay, and BS = batch size.}
\label{table:ModelParameters}
\end{table}

\paragraph{Clean Accuracy on the Combined dataset:}

To better understand overall clean accuracy, we also report per mark type performance on the Combined validation set. 

Tables~\ref{table:ResNet20CSubclass}--\ref{table:VGG16CSubclass} show the number of samples per mark type, the number of correct predictions, and the resulting accuracy for each Combined-trained model.

\begin{table}[h]
\centering
\small
\begin{tabular}{|c|c|c|c|}
\hline
\textbf{Mark Type} & \textbf{Total Samples} & \textbf{Correct Predictions} & \textbf{Accuracy} \\ \hline
Empty Bubbles & 5327 & 5327 & 100.00\% \\ \hline
Filled Bubbles & 1810 & 1810 & 100.00\% \\ \hline
Penrest & 703 & 703 & 100.00\% \\ \hline
Checkmark & 703 & 701 & 99.72\% \\ \hline
Crossmark & 703 & 675 & 96.02\% \\ \hline
Left right synthetic & 703 & 703 & 100.00\% \\ \hline
Back forth synthetic & 703 & 703 & 100.00\% \\ \hline
\end{tabular}
\caption{ResNet20-C Classification results for different mark types on the dataset.}
\label{table:ResNet20CSubclass}
\end{table}

\begin{table}[h]
\centering
\small
\begin{tabular}{|c|c|c|c|}
\hline
\textbf{Mark Type} & \textbf{Total Samples} & \textbf{Correct Predictions} & \textbf{Accuracy} \\ \hline
Empty Bubbles & 5327 & 5327 & 100.00\% \\ \hline
Filled Bubbles & 1810 & 1810 & 100.00\% \\ \hline
Penrest & 703 & 702 & 99.86\% \\ \hline
Checkmark & 703 & 701 & 99.72\% \\ \hline
Crossmark & 703 & 664 & 94.45\% \\ \hline
Left right synthetic & 703 & 703 & 100.00\% \\ \hline
Back forth synthetic & 703 & 703 & 100.00\% \\ \hline
\end{tabular}
\caption{CaiT-C classification results for different mark types on the dataset.}
\label{table:CaiTCSubclass}
\end{table}

\begin{table}[h]
\centering
\small
\begin{tabular}{|c|c|c|c|}
\hline
\textbf{Mark Type} & \textbf{Total Samples} & \textbf{Correct Predictions} & \textbf{Accuracy} \\ \hline
Empty Bubbles & 5327 & 5319 & 99.85\% \\ \hline
Filled Bubbles & 1810 & 1810 & 100.00\% \\ \hline
Penrest & 703 & 690 & 98.15\% \\ \hline
Checkmark & 703 & 665 & 94.59\% \\ \hline
Crossmark & 703 & 532 & 75.68\% \\ \hline
Left right synthetic & 703 & 703 & 100.00\% \\ \hline
Back forth synthetic & 703 & 703 & 100.00\% \\ \hline
\end{tabular}
\caption{SVM-C classification results for different mark types on the dataset.}
\label{table:SVMCSubclass}
\end{table}

\begin{table}[h]
\centering
\small
\begin{tabular}{|c|c|c|c|}
\hline
\textbf{Mark Type} & \textbf{Total Samples} & \textbf{Correct Predictions} & \textbf{Accuracy} \\ \hline
Empty Bubbles & 5327 & 5327 & 100.00\% \\ \hline
Filled Bubbles & 1810 & 1810 & 100.00\% \\ \hline
Penrest & 703 & 703 & 100.00\% \\ \hline
Checkmark & 703 & 701 & 99.72\% \\ \hline
Crossmark & 703 & 675 & 96.02\% \\ \hline
Left right synthetic & 703 & 703 & 100.00\% \\ \hline
Back forth synthetic & 703 & 703 & 100.00\% \\ \hline
\end{tabular}
\caption{VGG16-C classification results for different mark types on the dataset}
\label{table:VGG16CSubclass}
\end{table}

\section{Imperceivable Values}

\begin{figure*}
\begin{center}
\includegraphics[width=0.6\textwidth]{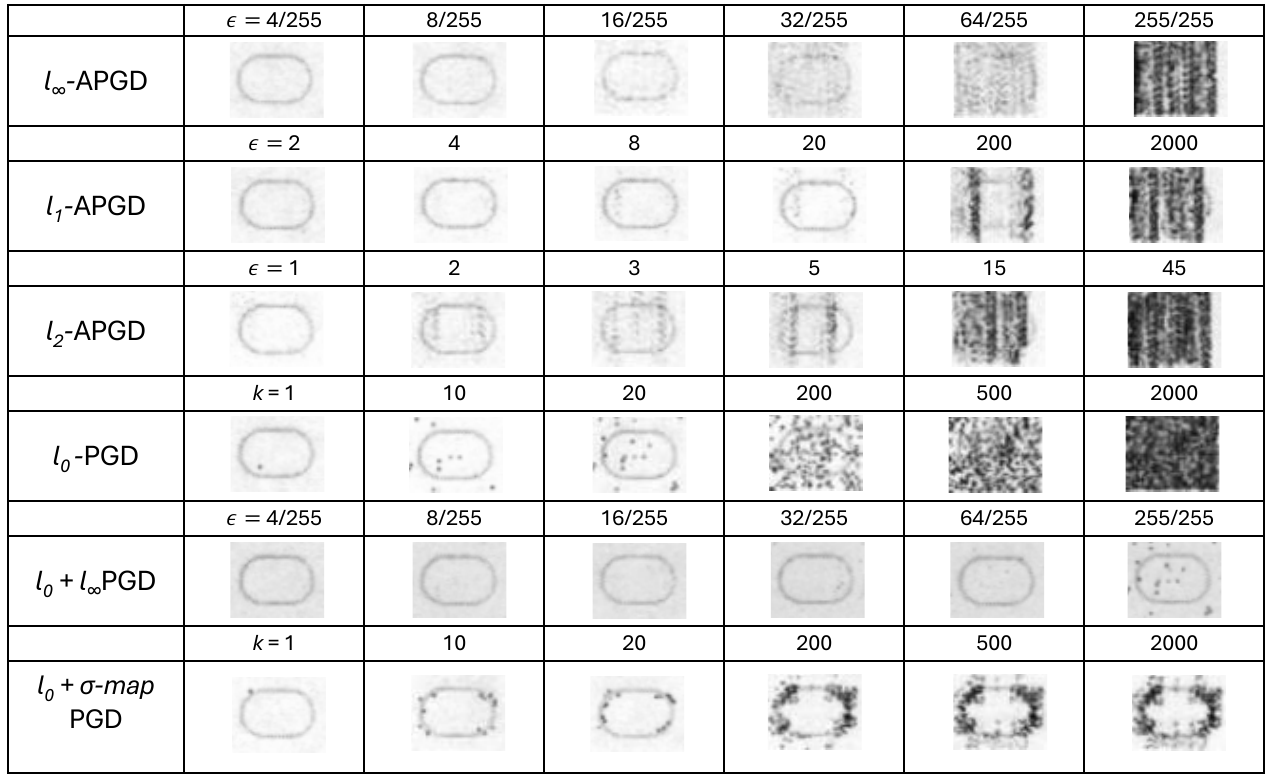} \end{center} \caption{Physical Adversarial examples with varying $\epsilon$ or $k$ for all attacks on the ResNet20-C model.} \label{fig:varyRModelFig} \end{figure*}

\begin{figure*}
\begin{center}
\includegraphics[width=0.6\textwidth]{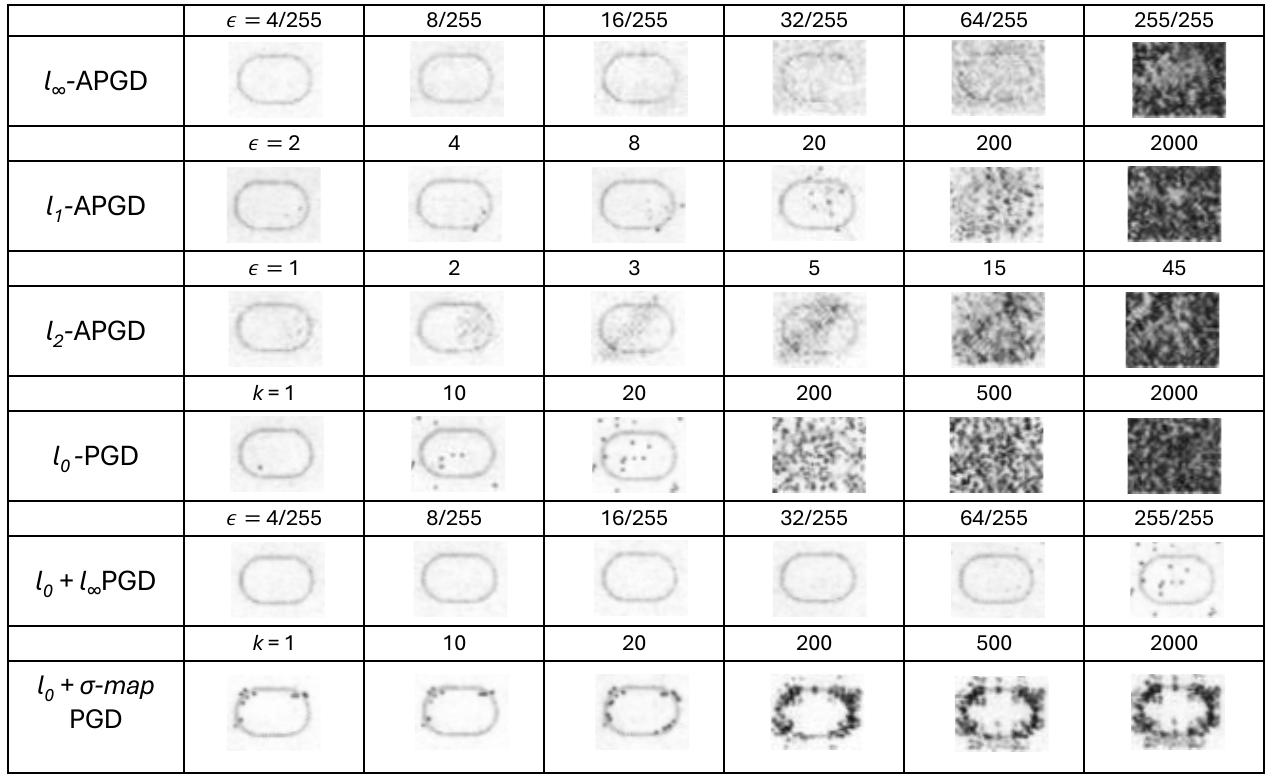} \end{center} \caption{Physical Adversarial examples with varying $\epsilon$ or $k$ for all attacks on the VGG16-C model.} \label{fig:varyVModelFig} \end{figure*}

\begin{figure*}
\begin{center}
\includegraphics[width=0.6\textwidth]{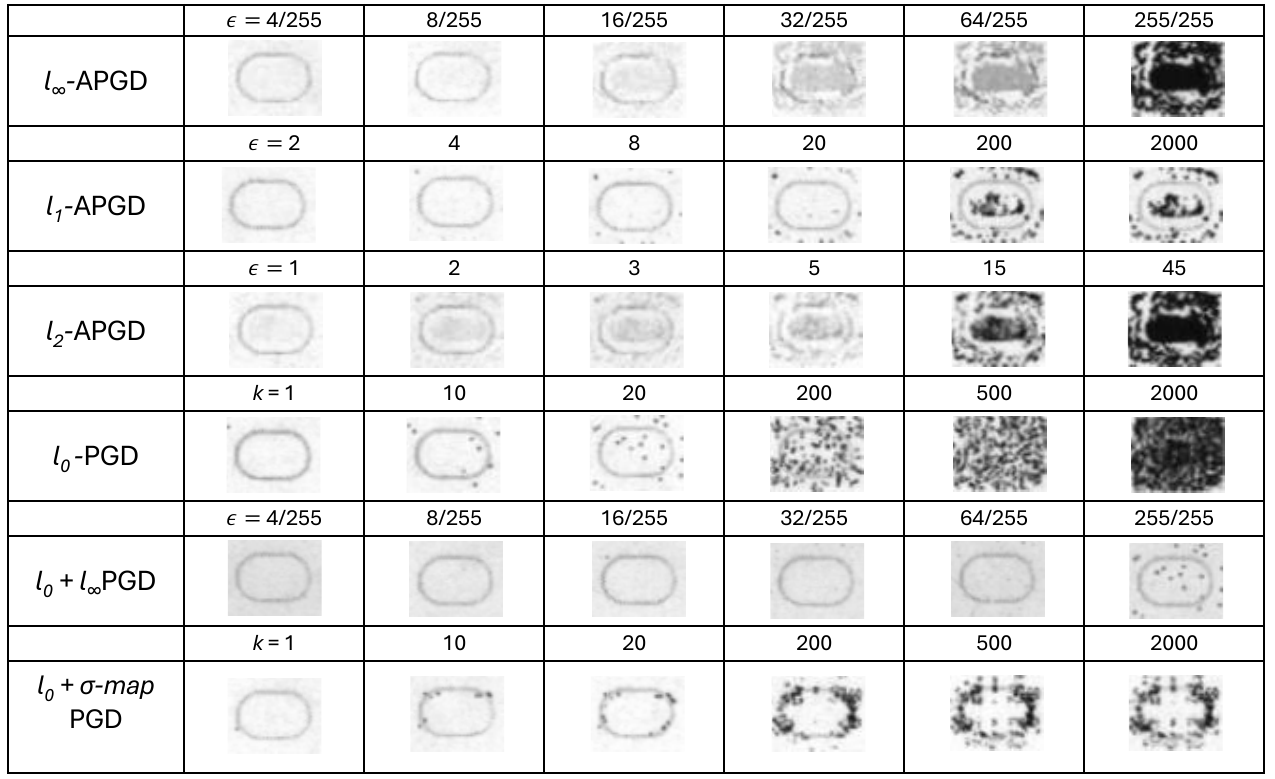} \end{center} \caption{Physical Adversarial examples with varying $\epsilon$ or $k$ for all attacks on the SVM-C model.} \label{fig:varySModelFig} \end{figure*}

We provide additional details for selecting imperceivable perturbation parameters used for both digital and physical experimental analyses.
Figures~\ref{fig:varyRModelFig}, \ref{fig:varyVModelFig}, and \ref{fig:varySModelFig} show physical adversarial examples with varying $\epsilon$ or $k$ for all six attacks on the ResNet20-C, VGG16-C, and SVM-C models, respectively. Along with the CaiT-C visualization in the main body (Figure~\ref{fig:varyModelFig}), these physical adversarial examples were provided to a panel of 6 U.S. election security experts to determine undetectable (imperceivable) noise levels for each attack.

Table~\ref{tab:physical-imperceptible-expert-votes} reports the individual perturbation values selected by each expert for each attack and model, along with summary statistics. For $l_{\infty}$-APGD and $l_{0}{+}l_{\infty}$ PGD, values denote the numerator of $\epsilon/255$. For $l_{1}$-APGD and $l_{2}$-APGD, values are $\epsilon$. For $l_{0}$-based attacks, values are $k$. We use the most common (overall) values (i.e., the mode) from Table~\ref{tab:physical-imperceptible-expert-votes} throughout the paper: $l_{\infty}$-APGD $\epsilon=16/255$, $l_{1}$-APGD $\epsilon=20$, $l_{2}$-APGD $\epsilon=2$, $l_{0}$ PGD $k=1$, $l_{0}{+}l_{\infty}$ PGD $\epsilon=64/255$, and $l_{0}{+}\sigma$-map PGD $k=1$. The corresponding average (overall) values in Table~\ref{tab:physical-imperceptible-expert-votes} are larger than the mode values, supporting the use of the most common values as undetectable noise levels across attacks.

\begin{table*}[t]
\centering
\small
\setlength{\tabcolsep}{4pt}
\renewcommand{\arraystretch}{1.15}
\begin{tabular}{|l|l|c|c|c|c|c|c|}
\hline
\textbf{Model} & \textbf{Expert} &
\textbf{$l_{\infty}$-APGD} &
\textbf{$l_{1}$-APGD} &
\textbf{$l_{2}$-APGD} &
\textbf{$l_{0}$ PGD} &
\textbf{$l_{0}{+}l_{\infty}$ PGD} &
\textbf{$l_{0}{+}\sigma$-map PGD} \\
\hline

\multirow{6}{*}{CaiT-C}
& Expert 1 & 4  & 4  & 1  & 1  & 32  & 1  \\ \cline{2-8}
& Expert 2 & 8  & 4  & 3  & 1  & 64  & 1  \\ \cline{2-8}
& Expert 3 & 4  & 4  & 4  & 1  & 64  & 1  \\ \cline{2-8}
& Expert 4 & 8  & 4  & 3  & 1  & 64  & 1  \\ \cline{2-8}
& Expert 5 & 4  & 4  & 2  & 1  & 64  & 10 \\ \cline{2-8}
& Expert 6 & 8  & 2  & 1  & 10 & 64  & 1  \\ \hline

\multirow{6}{*}{ResNet20-C}
& Expert 1 & 16 & 20 & 2  & 1  & 64  & 10 \\ \cline{2-8}
& Expert 2 & 32 & 20 & 2  & 10 & 64  & 10 \\ \cline{2-8}
& Expert 3 & 16 & 20 & 4 & --  & 64  & 1  \\ \cline{2-8}
& Expert 4 & 32 & 20 & 2  & 1  & 64  & 10 \\ \cline{2-8}
& Expert 5 & 16 & 20 & 2  & 10 & 64  & 10 \\ \cline{2-8}
& Expert 6 & 16 & 8  & 2  & 1  & 64  & 1  \\ \hline

\multirow{6}{*}{VGG16-C}
& Expert 1 & 32 & 20 & 3  & 10 & 64  & -- \\ \cline{2-8}
& Expert 2 & 32 & 8  & 2  & 10 & 255 & 1  \\ \cline{2-8}
& Expert 3 & 16 & 4 & 4  &  --  & 64 & --  \\ \cline{2-8}
& Expert 4 & 32 & 20 & 2  & 1  & 255 & 1  \\ \cline{2-8}
& Expert 5 & 16 & 8  & 2  & 10 & 64  & 10 \\ \cline{2-8}
& Expert 6 & 8  & 8  & 2  & 1  & 255 & 1  \\ \hline

\multirow{6}{*}{SVM-C}
& Expert 1 & 16 & 20 & 5  & 4  & 64  & 20 \\ \cline{2-8}
& Expert 2 & 16 & 20 & 3  & 10 & 255 & 200\\ \cline{2-8}
& Expert 3 & 16 & 20 & 4  & 1  & 64 & 1  \\ \cline{2-8}
& Expert 4 & 16 & 20 & 2  & 10 & 255 & 10 \\ \cline{2-8}
& Expert 5 & 8  & 8  & 3  & 10 & 64  & 20 \\ \cline{2-8}
& Expert 6 & 8  & 8  & 1  & 1  & 64  & 10 \\ \hline

\multicolumn{2}{|c|}{\textbf{Mode (Most Common)}} & \textbf{16} & \textbf{20} & \textbf{2} & \textbf{1} & \textbf{64} & \textbf{1} \\ \hline
\multicolumn{2}{|c|}{\textbf{Average}}     & 16.35 & 12.25 & 2.54 & 4.82 & 102.46 & 15.05 \\ \hline

\end{tabular}
\caption{Election security expert selected imperceptible values for each attack and model. For $l_{\infty}$-APGD and $l_{0}{+}l_{\infty}$ PGD, values denote the numerator of $\epsilon/255$. For $l_{1}$-APGD and $l_{2}$-APGD, values are $\epsilon$. For $l_{0}$-based attacks, values are $k$.}
\label{tab:physical-imperceptible-expert-votes}
\end{table*}
\section{Digital Attacks Details}

We provide additional details for the attack in virtual context. We describe how we selected attack hyperparameters, justify our choice of loss function, and report the complete digital robustness results for all six attacks across all models and budgets.

\paragraph{Attack Hyperparameter:}
For each attack, we first tested the largest budget (\(\epsilon\) or \(k\)) with a simple default setting. 
If the attack did not reach \(0\%\) robust accuracy, we increased the number of steps and/or adjusted the step size and/or changing random start until it succeeded. 
Once an attack reached \(0\%\) at the largest budget for a given model, we fix those optimization hyperparameters for that model and used the same settings for all smaller budgets. 
The final hyperparameters used for every attack and model are listed in Table~\ref{tab:apgd-linf-hparams}--\ref{tab:l0-sigma-hparams}. For \(l_{0}{+}l_{\infty}\) PGD we use the same hyperparameters as \(l_{0}\) PGD (Table~\ref{tab:l0-pgd-hparams}) with fixed sparsity \(k{=}20\).

\begin{table}[t]
\centering
\small
\setlength{\tabcolsep}{4pt}
\begin{tabular}{lccc}
\toprule
\textbf{Model} & \textbf{Step size init} & \textbf{\# Steps} & \textbf{Random start} \\
\midrule
SVM/VGG16/ResNet20 & $2\epsilon$ & 500  & False \\
CaiT                         & $0.05\epsilon$ & 1500 & False \\
\bottomrule
\end{tabular}
\caption{$l_{\infty}$-APGD hyperparameters.}
\label{tab:apgd-linf-hparams}
\end{table}

\begin{table}[t]
\centering
\small
\setlength{\tabcolsep}{4pt}
\begin{tabular}{lccc}
\toprule
\textbf{Model} & \textbf{Step size init} & \textbf{\# Steps} & \textbf{Random start} \\
\midrule
ResNet20-B &  $1.0\epsilon$ & 500  & False \\
ResNet20-C & $0.05\epsilon$ & 500  & False \\
CaiT-B     & $0.05\epsilon$ & 500  & False \\
CaiT-C     & $0.10\epsilon$ & 1250 & False \\
SVM-B      & $0.05\epsilon$ & 500  & False \\
SVM-C      & $0.10\epsilon$ & 500  & False \\
VGG16-B    & $0.10\epsilon$ & 500  & True \\
VGG16-C    & $0.10\epsilon$ & 500  & True \\

\bottomrule
\end{tabular}
\caption{\(l_{1}\)-APGD hyperparameters.}
\label{tab:apgd-l1-hparams}
\end{table}

\begin{table}[t]
\centering
\small
\setlength{\tabcolsep}{4pt}
\begin{tabular}{lccc}
\toprule
\textbf{Model} & \textbf{Step size init} & \textbf{\# Steps} & \textbf{Random start} \\
\midrule
ResNet20-B & $2\epsilon$ & 100 & False \\
ResNet20-C & $2\epsilon$ & 150 & False \\
CaiT-B     & $2\epsilon$ & 100 & False \\
CaiT-C     & $2\epsilon$ & 750 & False \\
SVM-B      & $2\epsilon$ & 100 & False \\
SVM-C      & $2\epsilon$ & 100 & False \\
VGG16-B    & $2\epsilon$ & 200 & False \\
VGG16-C    & $2\epsilon$ & 200 & False \\
\bottomrule
\end{tabular}
\caption{\(l_{2}\)-APGD hyperparameters.}
\label{tab:apgd-l2-hparams}
\end{table}

\begin{table}[t]
\centering
\small
\setlength{\tabcolsep}{4pt}
\begin{tabular}{lcccc}
\toprule
\textbf{Model} & \textbf{\# Steps} & \textbf{\# Step size} & \textbf{\# Restarts} & \textbf{Random start} \\
\midrule
ResNet20-B & 25 & 15 & 10 & True \\
ResNet20-C & 30 & 15 & 10 & True \\
CaiT-B     & 25 & 15 & 10 & True \\
CaiT-C     & 30 & 15 & 10 & True \\
SVM-B      & 25 & 15 & 10 & True \\
SVM-C      & 25 & 15 & 10 & True \\
VGG16-B    & 25 & 15 & 10 & True \\
VGG16-C    & 30 & 15 & 10 & True \\
\bottomrule
\end{tabular}
\caption{\(l_{0}\)-PGD hyperparameters.}
\label{tab:l0-pgd-hparams}
\end{table}

\begin{table}[t]
\centering
\small
\setlength{\tabcolsep}{4pt}
\begin{tabular}{lccccc}
\toprule
\textbf{Model} & \textbf{\# Steps} & \textbf{Step size} & \textbf{\(\kappa\)} & \textbf{\# Restarts} & \textbf{Random start} \\
\midrule
ResNet20-B & 75  & 15 & 10 & 10 & True \\
ResNet20-C & 75  & 15 & 10 & 10 & True \\
CaiT-B     & 500 & 15 & 30 & 1  & True \\
CaiT-C     & 500 & 15 & 30 & 1  & True \\
SVM-B      & 500 & 15 & 30 & 1  & True \\
SVM-C      & 75  & 15 & 10 & 10 & True \\
VGG16-B    & 75  & 15 & 10 & 10 & True \\
VGG16-C    & 75  & 15 & 10 & 10 & True \\
\bottomrule
\end{tabular}
\caption{\(l_{0}{+}\sigma\)-map PGD hyperparameters.}
\label{tab:l0-sigma-hparams}
\end{table}

\paragraph{Choice of Loss Function:}

In preliminary experiments we used APGD with cross-entropy (CE) loss at the largest budget for \(l_{\infty}\) (\(\epsilon=255/255\)) on listed trained models, it did not reach \(0\%\) robust accuracy on CaiT-C (Table~\ref{tab:ce_loss}), even after extensive tuning.

\begin{table}[t]
\centering
\small
\setlength{\tabcolsep}{6pt}
\begin{tabular}{lc}
\toprule
\textbf{Model} & \textbf{Robust acc. ($\epsilon{=}255/255$)} \\
\midrule
ResNet20-B & 0.000 \\
ResNet20-C & 0.000 \\
CaiT-B     & 0.000 \\
CaiT-C     & 0.017 \\
SVM-B      & 0.000 \\
SVM-C      & 0.000 \\
\bottomrule
\end{tabular}
\caption{Digital robust accuracy at the maximum budget for $l_{\infty}$-APGD using cross-entropy (CE) loss.}
\label{tab:ce_loss}
\end{table}
We therefore use the Difference of Logits Ratio (DLR) loss. Because our task is binary classification, we use the binary-safe variant from~\cite{Busting}, which drops the denominator. The modified DLR loss function is defined as:
\begin{equation}
\label{eq:DLR}
\text{DLR}(x, y) = -\bigl(z_y - \max_{j\neq y} z_{j}\bigr).
\end{equation}

\paragraph{Results}
Table~\ref{tab:apgd-linf-digital}--\ref{tab:l0-sigma-digital} reports the complete digital robustness results for all six attacks across all models and budgets. 

\begin{table*}[t]
\centering
\small
\setlength{\tabcolsep}{4pt}
\begin{tabular}{lcccccc}
\toprule
\textbf{Model} &
\textbf{$4/255$} &
\textbf{$8/255$} &
\textbf{$16/255$} &
\textbf{$32/255$} &
\textbf{$64/255$} &
\textbf{$255/255$} \\
\midrule
ResNet20-B & 1.000 & 0.493 & 0.219 & 0.000 & 0.000 & 0.000 \\
ResNet20-C & 0.404 & 0.002 & 0.000 & 0.000 & 0.000 & 0.000 \\
\hline
CaiT-B     & 1.000 & 1.000 & 1.000 & 1.000 & 0.614 & 0.000 \\
CaiT-C     & 0.983 & 0.501 & 0.500 & 0.498 & 0.492 & 0.000 \\
\hline
SVM-B      & 1.000 & 1.000 & 1.000 & 1.000 & 0.113 & 0.000 \\
SVM-C      & 0.500 & 0.500 & 0.500 & 0.385 & 0.000 & 0.000 \\
\hline
VGG16-B    & 0.978 & 0.302 & 0.000 & 0.000 & 0.000 & 0.000 \\
VGG16-C    & 0.914 & 0.257 & 0.000 & 0.000 & 0.000 & 0.000 \\
\bottomrule
\end{tabular}
\caption{Digital robust accuracy for \(l_{\infty}\)-APGD (DLR loss) at different \(\epsilon\) budgets.}
\label{tab:apgd-linf-digital}
\end{table*}

\begin{table*}[t]
\centering
\small
\setlength{\tabcolsep}{4pt}
\begin{tabular}{lcccccc}
\toprule
\textbf{Model} &
\textbf{$2$} &
\textbf{$4$} &
\textbf{$8$} &
\textbf{$20$} &
\textbf{$200$} &
\textbf{$2000$} \\
\midrule
ResNet20-B & 1.000 & 0.516 & 0.497 & 0.027 & 0.000 & 0.000 \\
ResNet20-C & 0.499 & 0.454 & 0.088 & 0.000 & 0.000 & 0.000 \\
\hline
CaiT-B     & 1.000 & 1.000 & 1.000 & 0.496 & 0.000 & 0.000 \\
CaiT-C     & 1.000 & 0.513 & 0.500 & 0.499 & 0.418 & 0.000 \\
\hline
SVM-B      & 1.000 & 1.000 & 1.000 & 1.000 & 0.000 & 0.000 \\
SVM-C      & 1.000 & 0.508 & 0.500 & 0.500 & 0.000 & 0.000 \\
\hline
VGG16-B    & 1.000 & 1.000 & 0.885 & 0.498 & 0.000 & 0.000 \\
VGG16-C    & 0.533 & 0.500 & 0.500 & 0.500 & 0.000 & 0.000 \\
\bottomrule
\end{tabular}
\caption{Digital robust accuracy for \(l_{1}\)-APGD (DLR loss) at different \(\epsilon\) budgets. }
\label{tab:apgd-l1-digital}
\end{table*}

\begin{table*}[t]
\centering
\small
\setlength{\tabcolsep}{4pt}
\begin{tabular}{lcccccc}
\toprule
\textbf{Model} &
\textbf{$\epsilon{=}1$} &
\textbf{$\epsilon{=}2$} &
\textbf{$\epsilon{=}3$} &
\textbf{$\epsilon{=}5$} &
\textbf{$\epsilon{=}15$} &
\textbf{$\epsilon{=}45$} \\
\midrule
ResNet20-B & 0.475 & 0.260 & 0.000 & 0.000 & 0.000 & 0.000 \\
ResNet20-C & 0.000 & 0.000 & 0.000 & 0.000 & 0.000 & 0.000 \\
\hline
CaiT-B     & 1.000 & 0.999 & 0.482 & 0.000 & 0.000 & 0.000 \\
CaiT-C     & 0.498 & 0.498 & 0.492 & 0.395 & 0.100 & 0.000 \\
\hline
SVM-B      & 1.000 & 1.000 & 1.000 & 0.991 & 0.000 & 0.000 \\
SVM-C      & 0.500 & 0.500 & 0.484 & 0.000 & 0.000 & 0.000 \\
\hline
VGG16-B    & 0.000 & 0.000 & 0.000 & 0.000 & 0.000 & 0.000 \\
VGG16-C    & 0.880 & 0.000 & 0.000 & 0.000 & 0.000 & 0.000 \\
\bottomrule
\end{tabular}
\caption{Digital robust accuracy for \(l_{2}\)-APGD (DLR loss) at different \(\epsilon\) budgets.}
\label{tab:apgd-l2-digital}
\end{table*}

\begin{table*}[t]
\centering
\small
\setlength{\tabcolsep}{4pt}
\begin{tabular}{lcccccc}
\toprule
\textbf{Model} &
\textbf{$k{=}1$} &
\textbf{$k{=}10$} &
\textbf{$k{=}20$} &
\textbf{$k{=}200$} &
\textbf{$k{=}500$} &
\textbf{$k{=}2000$} \\
\midrule
ResNet20-B & 1.000 & 0.998 & 0.871 & 0.500 & 0.383 & 0.000 \\
ResNet20-C & 0.998 & 0.500 & 0.500 & 0.500 & 0.500 & 0.000 \\
\hline
CaiT-B     & 1.000 & 1.000 & 1.000 & 1.000 & 0.655 & 0.000 \\
CaiT-C     & 1.000 & 0.998 & 0.952 & 0.500 & 0.493 & 0.000 \\
\hline
SVM-B      & 1.000 & 1.000 & 1.000 & 1.000 & 0.739 & 0.000 \\
SVM-C      & 1.000 & 0.910 & 0.590 & 0.489 & 0.000 & 0.000 \\
\hline
VGG16-B    & 1.000 & 0.515 & 0.500 & 0.500 & 0.500 & 0.000 \\
VGG16-C    & 0.889 & 0.500 & 0.500 & 0.500 & 0.500 & 0.000 \\
\bottomrule
\end{tabular}
\caption{Digital robust accuracy for \(l_{0}\) PGD (DLR loss) at different sparsity budgets \(k\).}
\label{tab:l0-pgd-digital}
\end{table*}

\begin{table*}[t]
\centering
\small
\setlength{\tabcolsep}{4pt}
\begin{tabular}{lcccccc}
\toprule
\textbf{Model} &
\textbf{$4/255$} &
\textbf{$8/255$} &
\textbf{$16/255$} &
\textbf{$32/255$} &
\textbf{$64/255$} &
\textbf{$255/255$} \\
\midrule
ResNet20-B & 1.000 & 1.000 & 1.000 & 1.000 & 1.000 & 0.871 \\
ResNet20-C & 1.000 & 0.955 & 0.527 & 0.501 & 0.500 & 0.500 \\
\hline
CaiT-B     & 1.000 & 1.000 & 1.000 & 1.000 & 1.000 & 1.000 \\
CaiT-C     & 1.000 & 1.000 & 1.000 & 1.000 & 0.997 & 0.952 \\
\hline
SVM-B      & 1.000 & 1.000 & 1.000 & 1.000 & 1.000 & 1.000 \\
SVM-C      & 1.000 & 1.000 & 1.000 & 1.000 & 1.000 & 0.590 \\
\hline
VGG16-B    & 1.000 & 1.000 & 1.000 & 1.000 & 1.000 & 0.500 \\
VGG16-C    & 1.000 & 1.000 & 1.000 & 0.995 & 0.537 & 0.500 \\
\bottomrule
\end{tabular}
\caption{Digital robust accuracy for \(l_{0}{+}l_{\infty}\) PGD (DLR loss) with fixed sparsity \(k{=}20\) at different \(\epsilon\) budgets.}
\label{tab:l0-linf-pgd-digital}
\end{table*}

\begin{table*}[t]
\centering
\small
\setlength{\tabcolsep}{4pt}
\begin{tabular}{lcccccc}
\toprule
\textbf{Model} &
\textbf{$k{=}1$} &
\textbf{$k{=}10$} &
\textbf{$k{=}20$} &
\textbf{$k{=}200$} &
\textbf{$k{=}500$} &
\textbf{$k{=}2000$} \\
\midrule
ResNet20-B & 1.000 & 1.000 & 1.000 & 0.500 & 0.456 & 0.000 \\
ResNet20-C & 0.558 & 0.500 & 0.500 & 0.500 & 0.500 & 0.000 \\
\hline
CaiT-B     & 1.000 & 1.000 & 1.000 & 1.000 & 0.901 & 0.477 \\
CaiT-C     & 0.930 & 0.624 & 0.535 & 0.507 & 0.425 & 0.000 \\
\hline
SVM-B      & 1.000 & 1.000 & 1.000 & 0.999 & 0.505 & 0.372 \\
SVM-C      & 1.000 & 0.999 & 0.990 & 0.496 & 0.352 & 0.000 \\
\hline
VGG16-B    & 0.774 & 0.573 & 0.500 & 0.500 & 0.500 & 0.000 \\
VGG16-C    & 0.500 & 0.500 & 0.500 & 0.500 & 0.500 & 0.000 \\
\bottomrule
\end{tabular}
\caption{Digital robust accuracy for \(l_{0}{+}\sigma\)-map PGD (DLR loss) at different sparsity budgets \(k\). }
\label{tab:l0-sigma-digital}
\end{table*}

\section{Physical World Additional Experimental Results}
We report robustness measurements for every attack at every perturbation value in the Table~\ref{tab:physical-apgd-linf}--\ref{tab:physical-l0-sigma}  after printing and scanning adversarial ballots. 
We include per-attack tables across all tested budgets for the Combined-trained models.

\begin{table*}
\centering
\small
\setlength{\tabcolsep}{4pt}
\begin{tabular}{lcccccc}
\toprule
\textbf{Model} &
\textbf{$4/255$} &
\textbf{$8/255$} &
\textbf{$16/255$} &
\textbf{$32/255$} &
\textbf{$64/255$} &
\textbf{$255/255$} \\
\midrule
ResNet20-C & 1.000 & 1.000 & 1.000 & 0.988 & 0.556 & 0.500 \\
CaiT-C     & 0.999 & 0.970 & 0.540 & 0.500 & 0.500 & 0.500 \\
SVM-C      & 1.000 & 1.000 & 0.583 & 0.500 & 0.121 & 0.000 \\
VGG16-C    & 1.000 & 1.000 & 1.000 & 0.960 & 0.607 & 0.500 \\
\bottomrule
\end{tabular}
\caption{Physical robust accuracy for \(l_{\infty}\)-APGD  on Combined-trained models at different \(\epsilon\) budgets.}
\label{tab:physical-apgd-linf}
\end{table*}

\begin{table*}[t]
\centering
\small
\setlength{\tabcolsep}{4pt}
\begin{tabular}{lcccccc}
\toprule
\textbf{Model} &
\textbf{$\epsilon{=}2$} &
\textbf{$\epsilon{=}4$} &
\textbf{$\epsilon{=}8$} &
\textbf{$\epsilon{=}20$} &
\textbf{$\epsilon{=}200$} &
\textbf{$\epsilon{=}2000$} \\
\midrule
ResNet20-C & 1.000 & 1.000 & 1.000 & 0.971 & 0.500 & 0.500 \\
CaiT-C     & 1.000 & 0.746 & 0.540 & 0.501 & 0.500 & 0.500 \\
SVM-C      & 1.000 & 1.000 & 0.963 & 0.603 & 0.005 & 0.003 \\
VGG16-C    & 0.999 & 0.993 & 0.944 & 0.708 & 0.500 & 0.500 \\
\bottomrule
\end{tabular}
\caption{Physical robust accuracy for \(l_{1}\)-APGD on Combined-trained models at different \(\epsilon\) budgets.}
\label{tab:physical-apgd-l1}
\end{table*}

\begin{table*}
\centering
\small
\setlength{\tabcolsep}{4pt}
\begin{tabular}{lcccccc}
\toprule
\textbf{Model} &
\textbf{$\epsilon{=}1$} &
\textbf{$\epsilon{=}2$} &
\textbf{$\epsilon{=}3$} &
\textbf{$\epsilon{=}5$} &
\textbf{$\epsilon{=}15$} &
\textbf{$\epsilon{=}45$} \\
\midrule
ResNet20-C & 1.000 & 0.989 & 0.816 & 0.510 & 0.500 & 0.500 \\
CaiT-C     & 0.813 & 0.503 & 0.500 & 0.466 & 0.317 & 0.198 \\
SVM-C      & 1.000 & 0.552 & 0.500 & 0.466 & 0.001 & 0.000 \\
VGG16-C    & 1.000 & 0.946 & 0.626 & 0.511 & 0.500 & 0.500 \\
\bottomrule
\end{tabular}
\caption{Physical robust accuracy for \(l_{2}\)-APGD on Combined-trained models at different \(\epsilon\) budgets.}
\label{tab:physical-apgd-l2}
\end{table*}

\begin{table*}
\centering
\small
\setlength{\tabcolsep}{4pt}
\begin{tabular}{lcccccc}
\toprule
\textbf{Model} &
\textbf{$k{=}1$} &
\textbf{$k{=}10$} &
\textbf{$k{=}20$} &
\textbf{$k{=}200$} &
\textbf{$k{=}500$} &
\textbf{$k{=}2000$} \\
\midrule
ResNet20-C & 1.000 & 0.899 & 0.644 & 0.500 & 0.500 & 0.500 \\
CaiT-C     & 1.000 & 0.995 & 0.938 & 0.501 & 0.500 & 0.500 \\
SVM-C      & 1.000 & 0.927 & 0.764 & 0.502 & 0.412 & 0.000 \\
VGG16-C    & 1.000 & 0.901 & 0.594 & 0.500 & 0.500 & 0.500 \\
\bottomrule
\end{tabular}
\caption{Physical robust accuracy for \(l_{0}\) PGD on Combined-trained models at different sparsity budgets \(k\). We use \(\epsilon=1\) for the per-pixel bound.}
\label{tab:physical-l0-pgd}
\end{table*}

\begin{table*}[t]
\centering
\small
\setlength{\tabcolsep}{4pt}
\begin{tabular}{lcccccc}
\toprule
\textbf{Model} &
\textbf{$4/255$} &
\textbf{$8/255$} &
\textbf{$16/255$} &
\textbf{$32/255$} &
\textbf{$64/255$} &
\textbf{$255/255$} \\
\midrule
ResNet20-C & 1.000 & 1.000 & 1.000 & 1.000 & 1.000 & 0.742 \\
CaiT-C     & 0.907 & 0.907 & 0.899 & 0.909 & 0.919 & 0.877 \\
SVM-C      & 0.992 & 0.988 & 0.967 & 0.960 & 0.960 & 0.833 \\
VGG16-C    & 1.000 & 1.000 & 1.000 & 1.000 & 1.000 & 0.645 \\
\bottomrule
\end{tabular}
\caption{Physical robust accuracy for \(l_{0}{+}l_{\infty}\) PGD on Combined-trained models at different \(\epsilon\) budgets (with fixed sparsity \(k{=}20\)).}
\label{tab:physical-l0-linf}
\end{table*}

\begin{table*}
\centering
\small
\setlength{\tabcolsep}{4pt}
\begin{tabular}{lcccccc}
\toprule
\textbf{Model} &
\textbf{$k{=}1$} &
\textbf{$k{=}10$} &
\textbf{$k{=}20$} &
\textbf{$k{=}200$} &
\textbf{$k{=}500$} &
\textbf{$k{=}2000$} \\
\midrule
ResNet20-C & 0.975 & 0.975 & 0.875 & 0.500 & 0.500 & 0.500 \\
CaiT-C     & 0.963 & 0.650 & 0.559 & 0.523 & 0.519 & 0.297 \\
SVM-C      & 1.000 & 1.000 & 1.000 & 0.664 & 0.506 & 0.188 \\
VGG16-C    & 0.999 & 0.999 & 0.952 & 0.504 & 0.500 & 0.500 \\
\bottomrule
\end{tabular}
\caption{Physical robust accuracy for \(l_{0}{+}\sigma\)-map PGD on Combined-trained models at different sparsity budgets \(k\).}
\label{tab:physical-l0-sigma}
\end{table*}


\end{document}